\RequirePackage{snapshot}

\documentclass[journal]{IEEEtran}
\usepackage{etoolbox}
\usepackage{ifpdf}

\usepackage{cite}

\usepackage[pdftex]{graphicx}
\graphicspath{{figures/}}
\DeclareGraphicsExtensions{.pdf,.jpeg,.png,.jpg}

\usepackage{amsmath,bm}
\usepackage{amssymb}
\usepackage{balance}

\usepackage{breqn}
\usepackage{algorithm,algorithmic}

\usepackage[caption=false,font=footnotesize]{subfig}

\usepackage{url}
\usepackage{empheq}

\usepackage{mfirstuc}

\usepackage{tabularx}
\usepackage{booktabs}
\usepackage{framed}
\usepackage{pdfpages}
\usepackage[percent]{overpic}
\usepackage{amsmath}

\usepackage{xcolor}

\newcommand{\revisionOne}[1]{#1}
\newcommand{\revisionOnePar}[1]{#1}
\newcommand{\revisionTwo}[1]{#1}
\newcommand{\revisionTwoPar}[1]{#1}
\usepackage[hidelinks]{hyperref}
\usepackage[capitalise]{cleveref}

\newcommand{\dsResults}[5]{
\begin{figure*}[tbhp]
\vspace{-1em}
		
\fourWCap{#1}{image-segmentation}{Image Segmentation}{ionosphere}{Ionosphere}{dermatology}{Dermatology}{madelon}{Madelon}%
\fourWCap{#1}{isolet}{Isolet}{mfatCombined}{MFEAT}{Yale}{Yale}{COIL20}{COIL20}%
\fourWCap{#1}{bioresponse}{Bioresponse}{gisette}{Gisette}{tumour}{Tumour}{ovarian}{Ovarian}%
\centering

\begin{minipage}[c]{#5\textwidth}
\caption{\normalsize #2}
\label{fig:#1}
		\end{minipage}	
	\hfill
\begin{minipage}[c]{#3\textwidth}
			{\includegraphics[width=\textwidth]{figures/#1-COIL20-RF-legend}}
		\end{minipage}	
\vspace{-1em}
\end{figure*}

}

\newcommand{\onecf}[2]{%
	\begin{figure*}[tbh]
				\vspace{-.5em}

		\centering
		\hfill
		\begin{minipage}[b]{.325\textwidth}
			{\includegraphics[width=\textwidth]{figures/#1-MedianGP-1-NL}\vspace{-.5em}}
		\end{minipage}
		\hfill
		\hfill
		\begin{minipage}[b]{.325\textwidth}  
			{\includegraphics[width=\textwidth]{figures/#1-BestGP-1-NL}\vspace{-.5em}}
		\end{minipage}
		\hfill
		\hfill
		\begin{minipage}[b]{.325\textwidth}  
			{\includegraphics[width=\textwidth]{figures/#1-LLE-1-NL}\vspace{-.5em}}
		\end{minipage} 
		\hfill \\
		\hfill
		\begin{minipage}[b]{.325\textwidth}
		\centering \hspace{.25cm} Median GP-MaL-LT
		\end{minipage}
		\hfill
		\hfill
		\begin{minipage}[b]{.325\textwidth}  
		\centering \hspace{.25cm} Best GP-MaL-LT
		\end{minipage}
		\hfill
		\hfill
		\begin{minipage}[b]{.325\textwidth}  
		\centering \hspace{.5cm} LLE
		\end{minipage} 
		\hfill \\
		\centering
		\caption{Visualisation of GP-MaL-LT vs LLE at one embedded dimension on the #2 dataset.}
		\label{fig:#1-1d-results}
		\vspace{-.5em}
	\end{figure*} 
}

\newcommand{\twocf}[2]{%
	\begin{figure*}[tbh]
				\vspace{-.5em}

		\centering
		\hfill
		\begin{minipage}[b]{.325\textwidth}
			{\includegraphics[width=\textwidth]{figures/#1-MedianGP-2}\vspace{-.5em}}
		\end{minipage}
		\hfill
		\hfill
		\begin{minipage}[b]{.325\textwidth}  
			{\includegraphics[width=\textwidth]{figures/#1-BestGP-2}\vspace{-.5em}}
		\end{minipage}
		\hfill
		\hfill
		\begin{minipage}[b]{.325\textwidth}  
			{\includegraphics[width=\textwidth]{figures/#1-LLE-2}\vspace{-.5em}}
		\end{minipage} 
		\hfill \\
		\hfill
		\begin{minipage}[b]{.325\textwidth}
		\centering \hspace{.25cm} Median GP-MaL-LT
		\end{minipage}
		\hfill
		\hfill
		\begin{minipage}[b]{.325\textwidth}  
		\centering \hspace{.25cm} Best GP-MaL-LT
		\end{minipage}
		\hfill
		\hfill
		\begin{minipage}[b]{.325\textwidth}  
		\centering \hspace{.5cm} LLE
		\end{minipage} 
		\hfill \\
		\centering
		\caption{Visualisation of GP-MaL-LT vs LLE at two embedded dimensions on the #2 dataset.}
		\label{fig:#1-2d-results}
		\vspace{-.5em}
	\end{figure*} 
}

\newcommand{\fourWCap}[9]{%
		\centering
		\hfill
		\begin{minipage}[b]{.24\textwidth}
			{\includegraphics[width=\textwidth]{figures/#1-#2-RF}}
		\end{minipage}
		\hfill
		\begin{minipage}[b]{.24\textwidth}  
			{\includegraphics[width=\textwidth]{figures/#1-#4-RF}}
		\end{minipage}
		\hfill
		\begin{minipage}[b]{.24\textwidth}  
			{\includegraphics[width=\textwidth]{figures/#1-#6-RF}}
		\end{minipage}
		\hfill
		\begin{minipage}[b]{.24\textwidth}  
			\centering
			{\centering \includegraphics[width=\textwidth]{figures/#1-#8-RF}}
		\end{minipage} \hfill %
}

\makeatletter
\newcommand\thefontsize[1]{{#1 The current font size is: \f@size pt\par}}
\makeatother

\begin{document}
%
\title{Genetic Programming for Manifold Learning:\\ Preserving Local Topology}
%
%
%

\author{Andrew~Lensen,~\IEEEmembership{Member,~IEEE,}
	Bing~Xue,~\IEEEmembership{Senior Member,~IEEE,}
	and~Mengjie~Zhang,~\IEEEmembership{Fellow,~IEEE}
	\thanks{This work was supported in part by the Marsden Fund of the New Zealand Government under Contracts VUW1913 and VUW1914 and the University Research Fund at Te Herenga Waka--Victoria University of Wellington under grant number 226161/4164.}%
	\thanks{The authors are with the Evolutionary Computation Research Group (ECRG), Victoria University of Wellington, Wellington 6140, New Zealand (e-mail: andrew.lensen@ecs.vuw.ac.nz; bing.xue@ecs.vuw.ac.nz; mengjie.zhang@ecs.vuw.ac.nz).}%
}%

\markboth{Journal of \LaTeX\ Class Files,~Vol.~14, No.~8, August~2015}%
{Shell \MakeLowercase{\textit{et al.}}: Bare Demo of IEEEtran.cls for IEEE Journals}

\maketitle

\begin{abstract}
Manifold learning methods are an invaluable tool in today's world of increasingly huge datasets. Manifold learning algorithms can discover a much lower-dimensional representation (embedding) of a high-dimensional dataset through non-linear transformations that preserve the most important structure of the original data. State-of-the-art manifold learning methods directly optimise an embedding without mapping between the original space and the discovered embedded space. This makes interpretability --- a key requirement in exploratory data analysis --- nearly impossible. Recently, genetic programming has emerged as a very promising approach to manifold learning by evolving functional mappings from the original space to an embedding. However, genetic programming-based manifold learning has struggled to match the performance of other approaches. In this work, we propose a new approach to using genetic programming for manifold learning, which preserves local topology. This is expected to significantly improve performance on tasks where local neighbourhood structure (topology) is paramount. We compare our proposed approach with various baseline manifold learning methods and find that it often outperforms other methods, including a clear improvement over previous genetic programming approaches. These results are particularly promising, given the potential interpretability and reusability of the evolved mappings.
\end{abstract}

\begin{IEEEkeywords}
Manifold Learning, Genetic Programming, Feature Selection, Feature Construction, Evolutionary Learning.	\vspace{-.5em}
\end{IEEEkeywords}

\IEEEpeerreviewmaketitle

\makeatletter
\setlength{\@fptop}{0pt}
\setlength{\@fpbot}{0pt plus 1fil}
\makeatother

\section{Introduction}

\IEEEPARstart{M}{anifold} learning (MaL) techniques can greatly reduce the dimensionality of data by discovering a low-dimensional, non-linear \textit{embedding}, which retains the main manifold structure from the high-dimensional (input) data \cite{lee2007nonlinear} in a much smaller space. Such techniques are also commonly referred to as non-linear dimensionality reduction algorithms, due to their applications in tasks such as feature extraction \cite{bengio2013representation} and visualisation \cite{espadoto2019towards}.

A common categorisation of MaL methods is by how a method maps the high- to low-dimensional spaces \cite{lee2007nonlinear}. \textit{Explicit} mapping methods, which directly optimise each data point's position in the low-dimensional space, are by far the most prevalent in recent years \cite{maatenTSNE,2018arXivUMAP}. However, such approaches cannot be easily applied (generalised) to new data points and appear as ``black boxes'' as there is no functional relationship between the high- and low-dimensional spaces. In contrast, \textit{implicit} mapping methods produce a \textit{functional} mapping, which directly transforms the input space to a lower-dimensional one by finding a mathematical transformation. Such methods are easily reusable and are intrinsically more interpretable, as the transformation itself can be examined directly. However, learning such a function that can preserve the manifold structure as successfully as an explicit mapping method is challenging.

An inherent trade-off must be optimised by MaL methods: the degree to which the \textit{local} versus the \textit{global} topology of the input data is preserved in the embedded space \cite{espadoto2019towards}. Optimising local topology preserves the \textit{neighbourhood} of each instance, whereas global topology optimisation will better retain the overall dataset structure (e.g.\ any natural clusters). On most non-toy datasets, it is impossible to perfectly preserve both the global and local topologies in an embedding due to the much lower dimensionality of the embedded space. MaL methods have been proposed that optimise the preservation of either local \cite{roweis2000lle,tenenbaum2000isomap} or global \cite{kruskal1964mds,jolliffe2011pca} structure; or some combination of both \cite{meng2011new,cunningham2015linear}.

Genetic Programming (GP) \cite{poli2008field}, an Evolutionary Computation (EC) method, has recently been proposed for performing MaL with an implicit mapping approach, by utilising a functional tree-based representation \cite{lensen2019can,orzechowski2020benchmarking,lensen2020genetic}, which produces a transformation in the form of lisp-style program trees. These works showed promising results, but highlighted the need for further improvements in performance, especially when compared to state-of-the-art explicit mapping methods such as t-distributed Stochastic Neighbourhood Embedding (t-SNE) \cite{maatenTSNE} or Uniform Manifold Approximation and Projection for Dimension Reduction (UMAP) \cite{2018arXivUMAP}. In particular, the fitness functions prioritised global structure preservation at the expense of local neighbourhood topology.

Preserving local topology is paramount in many tasks where the relationship between similar instances is important. For example, in classification or image segmentation tasks, prioritising global structure at the expense of local topology will cause misclassification of instances by introducing overlap between distinct classes or erroneous movement of instances to different classes altogether. This is also problematic in exploratory data analysis tasks such as visualisation, where the user may be misled into believing two instances are unrelated if the local topology is not adequately preserved. \revisionOne{Consider the example of a medical doctor using machine learning to make medical predictions. They want a very accurate and efficient classifier and so may want a MaL method to reduce the dimensionality of their very high-dimensional medical data. But, they also need an interpretable model --- otherwise, how can they trust the model with their patients' well-being? For a model to be interpretable, the features it uses as inputs must also be interpretable. Existing MaL methods do not produce interpretable embeddings (with respect to the original feature space) and thus will cause the model to be uninterpretable if they are used. GP-based MaL, on the other hand, has the potential to effectively reduce dimensionality by using an interpretable mapping.}

In this paper, we propose the first approach to performing GP for MaL that focuses on preserving local topology. This is expected to greatly improve the performance on tasks that rely on local structure while also reinforcing the value of a GP-based MaL approach in producing an \textit{implicit} mapping. The major contributions of this paper are:

\begin{itemize}
	\item The formulation of a fitness function to measure how well local topology is preserved by the evolved mapping (represented by GP trees);
	\item The introduction of a surrogate model to approximate nearest-neighbours, greatly reducing computational cost while still ensuring the preservation of local structure;
	\item Rigorous experimental evaluation of the proposed method which shows clear benefit and improved performance against five baseline MaL methods over \revisionTwo{12} datasets;
	\item Extensive sensitivity analysis of the surrogate model which reinforces its robustness and provides guidance on parameter settings;
	\item Further analysis of the embeddings produced by the proposed method to understand how it performs compared to the canonical Locally Linear Embedding (LLE)\cite{roweis2000lle} method, which also focuses on preserving local topology.
\end{itemize}

\section{Background}

\subsection{Dimensionality Reduction}
Dimensionality reduction (DR) techniques are frequently used to improve the efficiency and interpretability of various machine learning tasks by transforming the input space into a much lower-dimensional one. These methods are loosely grouped into feature selection (FS; choosing a smaller subset of the original space) and feature construction (FC; combining features into new high-level features) \cite{liu2012feature}. EC methods have seen significant success in performing FS and FC \cite{xue2015survey,alsahaf2019survey}. 

Tree-based GP, in particular, has been applied to a wide range of FC problems, as its ability to automatically learn a symbolic model (mapping) from a set of inputs to an output is a natural representation for performing feature construction \cite{espejo2010survey,neshatian2012filter,lensen2017GPGC}. Its use in unsupervised FC, however, is notably lacking.

\subsection{Manifold Learning}
Manifold learning is an unsupervised form of non-linear dimensionality reduction that aims to find a mapping $f:x \to y$, where $x$ is a given feature space and $y$ is a much lower-dimensional embedded space that retains as much structure from $x$ as possible (e.g.\ minimising $Cost(y|x)$). In practice, most state-of-the-art MaL methods do not search for a function $f$ but instead directly optimise the $y$-space. These ``explicit" mapping methods are, in some sense, easier to develop, as any numerical search method can be used to optimise the $n\times d$-dimensional embedded space for $n$ instances and an embedded dimensionality of $d$. ``Implicit" mapping methods, which instead search for $f$ itself (i.e.\ perform FC), are understudied, likely due to the difficulty of representing and searching the space of non-linear functions. GP has been successfully used for related problems such as symbolic regression, and initial work into using GP for MaL has shown promise \cite{lensen2019multi,orzechowski2020benchmarking}.

\subsection{Measuring the Preservation of Local Topology}
The notation of a \textit{neighbourhood} is key to capturing the local topology of data. Simply put, each instance's neighbourhood in the embedded space should be the same as in the original input space. The canonical definition of a neighbourhood considers the $k$ closest instances (by some similarity measure) as the neighbours of a particular instance \cite{eppstein1997nn}. The quality of an embedding can then be evaluated by comparing each instance's input and embedded neighbourhoods --- if each of the neighbourhoods contains the same instances, with the same ordering, it is an optimal embedding.

This concept can be formulated in a variety of ways \cite{espadoto2019towards,meng2011new}. A straightforward approach is to simply count the number of matching neighbours for a dataset \cite{chen2009local}:  
\begin{equation}
\text{Local Continuity} = \sum_{i \in \text{Dataset}} |N(x_i) \cup N(y_i)|
\end{equation}
where $x_i$ is the $i^{th}$ instance in the input space with representation $y_i$ in the embedded space and $N(a)$ is the set of instances forming the neighbourhood of $a$.

However, such an approach fails to consider the \textit{ordering} of the nearest neighbours. For example, if the third nearest-neighbour in the input space becomes the seventh nearest-neighbour in the embedded space, this is clearly a worse result than if it had remained in the same position. To address this limitation, the local continuity measure can be extended to consider the \textit{rank} of neighbours in each space. Many measures based on neighbour ranks have been proposed, including the Trustworthiness \& Continuity (T\&C) measure \cite{venna2006local}:
\begin{align}
\text{Trustworthiness} = 1 - H_K \sum_{i \in D} \sum_{j \in U_K(i)} r(i,j) - K \\
\text{Continuity} = 1 - H_K \sum_{i \in D} \sum_{j \in V_K(i)} \hat{r}(i,j) - K
\end{align}
where $U_K(i)$ is the set of instances in the neighbourhood of $i$ in the \textit{embedded} space but not in the input space and $V_K(i)$ is the set of instances in the neighbourhood of $i$ in the \textit{input} space but not in the embedded space. $r(i,j)$ is the rank of neighbour $j$ in the input space; $\hat{r}(i,j)$ is the rank of neighbour $j$ in the embedded space. $H_k$ is a normalisation term based on the number of instances ($n$) and neighbours ($K$):
\begin{equation}
H_K = \frac{2}{nK(2n-3K-1)}
\end{equation}
By considering the rank of incorrect neighbours, the T\&C measure gives a more granular measure of local topology preservation than continuity alone. The T\&C measure can either consider trustworthiness and continuity separately or as a weighted scalar objective using a weighting parameter $\lambda$:

\begin{equation}
\text{T\&C} = (1-\lambda) \times \text{Trustworthiness} + \lambda \times \text{Continuity}
\end{equation}

Other measures consider how well a neighbour's \textit{distance} is maintained from the input to the embedded space, e.g.\, by calculating normalised error \cite{martins2014visual} or to what extent distances are \textit{stretched} or \textit{compressed} \cite{aupetit2007visualizing}. These approaches are even more granular than ranking methods, but they suffer from the ``curse of dimensionality'' on high-dimensional datasets, where distances become increasingly similar. In this paper, we consider that the \textit{ordering} of neighbours is sufficient for capturing the structure of data, especially in high-dimensional datasets.

\subsection{Related Work}
There are a handful of closely related works which use GP for MaL \cite{lensen2019can,lensen2019multi,lensen2020genetic,orzechowski2020benchmarking}. The earliest method, GP-MaL \cite{lensen2019can}, proposed using a multi-tree GP structure and a fitness function that attempts to measure both local and global topology preservation simultaneously. While GP-MaL demonstrated clear potential, the classification accuracy compared to other MaL approaches was lacking. GP-MaL is the existing GP baseline for this work. 

Evolutionary multi-objective GP-based MaL methods have also been proposed. The first work, GP-MaL-MO \cite{lensen2019multi}, explored the trade-off between dimensionality and embedding quality. A reduction in dimensionality of up to $95\%$ was attained without any significant loss of manifold structure. Another paper proposed the use of GP-based MaL for evolving \textit{interpretable} visualisations \cite{lensen2020genetic}. This GP-tSNE method showed that it was possible to produce high-quality visualisations using a GP-evolved function that could be interpreted and understood by humans. 

Orzechowski et al.\ \cite{orzechowski2020benchmarking} proposed ManiGP, a GP-based MaL approach that optimises the separability of classes in the embedded space. Their proposed fitness function considers the classification accuracy obtained when performing $k$-means clustering on the embedded space, and so ManiGP is not strictly an unsupervised learning method as it uses class labels to feedback into the optimisation process. The authors also provided a comprehensive benchmark of MaL methods (including GP-MaL and ManiGP) across a wide range of datasets.

There are also a few non-EC methods that attempt to perform MaL based on finding an explicit mapping. Autoencoders \cite{hinton2006reducing} are one such approach, but they tend to use a vast number of layers and weights, which limits their interpretability potential. They are also restricted to the use of differential operators and activation functions, which limits the types of feature relationships they can effectively capture. Parametric t-SNE \cite{maaten2009learning} is an extension of the canonical t-{SNE} algorithm that uses a neural network to learn a mapping from the original to the embedded space. This method, like the autoencoder, uses a huge number of neurons (over 10,000), which also limits its usefulness in practice.

In summary, many existing methods use explicit mappings, which allows them to effectively preserve topology in the embedded space at the expense of interpretability, reusability, and generalisability. GP-based MaL methods, which do not have these limitations, have shown promising results, but their performance is still worse than explicit mapping methods. We expect that using a local topology preservation approach will make significant progress in improving GP-based MaL's performance.

\section{The Proposed Method: GP-MaL-LT}
In this work, we propose a new Genetic Programming for Manifold Learning method that preserves Local Topology (GP-MaL-LT). A multi-tree GP representation is used, where each individual comprises of $d$ GP trees. Each tree represents a single dimension in the embedding. The number of trees should be set \textit{a priori}, as in the majority of MaL algorithms. The fitness of a GP individual is based on the quality of the embedding produced by evaluating the output of all trees to get a $d-dimensional$ embedding. The remainder of the GP algorithm is essentially unchanged: a population of individuals is randomly initialised and then repetitively evaluated and updated for a number of generations, producing one final solution (the individual with the best fitness).

The following subsections describe the main aspects of the proposed approach in turn: first, the GP representation, then the formulation of the novel fitness function, and finally, the computationally efficient surrogate model. 

\subsection{Genetic Programming Representation}
The function set used in previous work \cite{lensen2019can,lensen2019multi} was shown to perform well in manifold learning tasks. It contains a range of functions that allow for mapping a variety of complex relationships in a local topology, and so we use a similar representation in this work. This is summarised in \cref{terminalAndFunctionSet}. A variety of arithmetic, non-linear, and conditional functions allow for a wide variety of non-linear mappings to be found to represent as much of the local topology as possible in the embedded space. The terminal set contains the features of the dataset (i.e.\ the input dimensions), as well as an ephemeral random constant (ERC) which allows for weighting within a tree. \revisionOne{All ERCs are in the range $[-1,1]$.}
\begin{table}
	\vspace{-1em}
	\centering
	\caption{The function and terminal sets of the proposed method.}
	\vspace{-.5em}
	\label{terminalAndFunctionSet}
	\begin{tabularx}{.7\linewidth}{@{}Xcr@{}}
		\toprule
		Function & No.\ of Inputs & Description\\
		\midrule
		\multicolumn{3}{c}{Arithmetic Functions} \\
		\midrule
		$+$ & 2 & Addition\\
		$-$ & 2 & Subtraction\\
		$\times$ & 2 & Multiplication\\
		$\div$ & 2 & Protected Division\\
		\midrule
		\multicolumn{3}{c}{Non-Linear Functions}\\
		\midrule
		Sigmoid & 1 & $\frac{1}{1+e^{-x}}$\\
		ReLU &1&$\max(0,x)$\\
		\midrule
		\multicolumn{3}{c}{Conditional Functions}\\
		\midrule
		Max & 2 & $\max(x,y)$\\
		Min & 2 & $\min(x,y)$\\
		If & 3 & if $(x< 0)$: y; else $z$\\
		\midrule
		\multicolumn{3}{c}{Terminal Nodes}\\
		\midrule
		F$_i$ & 0 & $i^{th}$ feature value\\
		Constant & 0 & From $U[-1,1]$\\
		\bottomrule
	\end{tabularx}
	\vspace{-1em}
\end{table}

Crossover is performed using the ``all-pairs'' multi-tree approach as in existing work \cite{lensen2019multi}, where each pair of matching trees in the two-parent individuals are crossed over to produce two offspring. Mutation is performed by selecting a single random tree from the parent and then performing mutation as standard on that tree. 

\subsection{Formulation of Fitness Function}
In this work, we propose a ranking-based cost function to measure local topology preservation, with several adjustments to produce a smoother fitness space, reduce the computational cost, and remove the need for a trade-off $\lambda$ free parameter. 

Using a ranked-based approach as a fitness function in any metaheuristic (e.g.\ GP) presents an immediate challenge: ranking neighbours for an individual requires computing and sorting all pairwise distances in the embedded space, at a cost of $\Theta(n^2)$ + $\Theta(n \log{n})$ for $n$ instances in the dataset. For a population of size $P$ evolving over $G$ generations, this gives a complexity for fitness evaluation alone of $O(PGn^2)$, which quickly becomes unrealistic for reasonably large datasets. To remedy this, we utilise a nearest-neighbour \textit{approximation}/surrogate technique (detailed further in \cref{ann}), which does not require a full pairwise distance matrix to be computed. 

In the case where a neighbour correctly appears in both the K-neighbourhood in both the input and embedded spaces, we can simply calculate how much it \textit{deviates} in the embedded ranking from the correct position in the input ranking as follows:
\begin{equation}
Deviation(i,j) = \frac{|r(i,j) - \hat{r}(i,j) |}{\max\{r(i,j), K-r(i,j)\}}
\end{equation}
where $i$ is the instance being considered, $j$ is the $j^{th}$ nearest neighbour, $r(i,j)$ is the rank of $j$ in the input space, and  $\hat{r}(i,j)$ is the rank of $j$ in the embedded space. The numerator of this formula calculates how far $j$ deviates in the embedded space from its correct position in the input space, and the denominator is a normalisation term that ensures each neighbour has a deviation between $0$ (no deviation; exactly the same position) and $1$ (maximal deviation; opposite position). 

Based on this measure of deviation, we propose a cost function to measure the local cost of $i$, defined as:
\begin{equation}
\label{costFunc}
Cost(i) = |V_k(i)| + \frac{1}{|W_k(i)|}\sum_{j \in W_K(i)}Deviation(i,j)
\end{equation}
where $V_k(i)$ is the set of instances in the neighbourhood of $i$ in the input space but \textbf{not} in the embedded space, and $W_k(i)$ is the set of instances in the neighbourhood of $i$ in \textbf{both} the input and embedded spaces. This cost function consists of two parts. The first part simply counts the number of missing neighbours in the embedded space\footnote{A neighbour may be in the $k$-neighbourhood in the high-dimensional space, but not in the embedded one.}, i.e.\ $|V_k(i)|$. This is bounded between $0$ and $K$, where a smaller value indicates fewer missing neighbours. The second part of this cost function considers the deviation of those instances that are correctly in the neighbourhood in both spaces (those in $W_k(i)$). This second term is effectively the average deviation, bounded between $0$ and $1$. By formulating the cost function in this way, it ensures that \textit{missing} neighbours always give a greater penalty than \textit{misordered} neighbours. This is an important observation: it is much worse for a neighbour not to appear in an embedding's neighbourhood than for it to appear in the wrong order. 

Until now, we have not considered the significance of the ordering of neighbours in the input neighbourhood. The closer a neighbour in the input space, the more important it is to the local topology of an instance --- that is, the $n^{th}$-nearest-neighbour is fundamentally more local (more important) than the ${n+1}^{th}$ neighbour. To capture this relationship, we propose weighting the cost function by how near each neighbour is in the instance's neighbourhood:
\begin{dmath}
Cost_{Weighted}(i) = \sum_{j \in V_k(i)} \frac{K-j}{K} + \frac{1}{|W_k(i)|}\sum_{j \in W_K(i)}\frac{K-j}{K}Deviation(i,j)
\end{dmath}
This equation is very similar to \cref{costFunc}, with the main change being the addition of the two $\frac{K-j}{K}$ terms, which weigh each neighbour between $1$ and $0$ based on how close they are to the instance being considered.

The fitness of a given individual, $Ind$, is then simply the mean weighted cost across all instances in the dataset:
\begin{equation}
Fitness(Ind) = \frac{1}{n} \sum_{i \in \text{Dataset}} Cost_{weighted}(i)
\end{equation}
This fitness should be minimised and is in the range $[0, K]$ --- in the worst case, all $K$ neighbours for all $n$ instances will be missing from the embedded neighbourhood.
\subsection{Approximating Nearest-Neighbours}
\label{ann}
\revisionOnePar{
As mentioned earlier, there is a key limitation in all ranking-based cost measures: computing ranks requires sorting the full pairwise distance matrix, at $\Theta(n^2)$. This is very expensive when it must be done for every evaluation of a GP individual. To remedy this, we use a surrogate approach to \textit{approximate} the nearest-neighbours of an instance in the embedded space. We choose neighbours based on Euclidean distance, as this is the predominant approach in the MaL literature. 

In this work, we use the Hierarchical Navigable Small World (HNSW) algorithm \cite{yury2020efficient,aumuller2020ann}. Specifically, we use the hnswlib library\footnote{\url{https://github.com/nmslib/hnswlib}}. HNSW can approximate a $K$-neighbourhood for a given GP individual in $O(Kn \log{n})$. Assuming that $K \ll n$ (for non-trivial data), this gives a net complexity of $O(n \log{n})$, which is clearly much more feasible than the $\Theta(n^2)$ for the exact (pairwise) method. 
}


\section{Experiment Design}
It is very difficult to fairly compare MaL algorithms in an unsupervised manner (e.g.\ using a formulation of cost/quality) as this would inevitably bias the comparison towards methods that most closely optimise that formulation. Thus, one of the most common approaches to quantitatively compare the performance of MaL algorithms is to compare the attainable classification accuracy on the embedded space produced by each algorithm. Theoretically, other supervised learning tasks such as regression could also be used as a proxy for measuring MaL quality. However, classification tasks are especially appropriate for evaluating how well \textit{local} topology has been preserved in the embeddings, as neighbouring instances are expected to be members of the same (sub-)class.

In this work, we focus our evaluation on the test accuracy achieved when training a classifier on the embeddings produced by the proposed method and the baseline MaL methods. To reduce variability caused by the classifier itself, we use ten-fold cross-validation with a random forest classifier (RF) using an ensemble of $100$ trees. Large ensembles of RFs are known to give excellent and stable performance, making them appropriate as an evaluation technique.

\revisionOnePar{It is important to note that, as we use both implicit and explicit mapping methods as baselines, we can only split the datasets into training and test sets \textit{after} performing manifold learning on the whole dataset. Explicit mapping methods cannot be applied to out-of-sample data (i.e.\ new examples) without re-optimisation. Given the lack of implicit mapping methods (particularly ones that optimise preservation of local topology) in the literature, we include several explicit mapping methods in our evaluation to give a clearer picture of the performance of the proposed method. }

\subsection{Baseline Methods}
The five baseline methods were selected from a variety of manifold learning paradigms and include:

\begin{itemize}
	\item Genetic Programming for Manifold Learning (GP-MaL)\cite{lensen2019can}: the first attempt at applying GP to MaL, which uses a fitness function that does not differentiate between preserving local and global topology.
	\item Principal Component Analysis (PCA)\cite{jolliffe2011pca}: computes a number of orthogonal linear transformations of the data, where each successive axis captures the direction of the maximum remaining variance. While technically not a MaL method (as it does not use non-linear transforms), PCA provides a useful baseline of what is attainable with linear transformations alone.
	\item Locally Linear Embedding (LLE)\cite{roweis2000lle}: tries to preserve local structure by modelling each instance as a weighted combination of its neighbours and then attempting to maintain this weighting in the embedded space. LLE is a canonical example of a MaL method that optimises local topology preservation.
	\item MultiDimensional Scaling (MDS) \cite{kruskal1964mds,saeed2018survey}: a well-known example of a MaL method which attempts to preserve the \textit{distance} between neighbours in the input and embedded spaces. 
	\item Uniform Manifold Approximation and Projection for Dimension Reduction (UMAP) \cite{2018arXivUMAP}: generally regarded as the state-of-the-art MaL method in terms of raw structure-preservation performance; uses a fuzzy topological structure to model the input and embedded spaces. Often used instead of the canonical t-Distributed Stochastic Neighbour Embedding (t-SNE) \cite{maatenTSNE} method nowadays.
\end{itemize}

Of the above approaches, LLE, MDS, and t-SNE are \textit{explicit} mapping methods, which directly optimise instances' positions in the embedded space. Only GP-MaL and PCA are \textit{implicit} mapping methods, producing a parametrised mapping from the input to embedded spaces. 

MaL methods either require the number of dimensions in the embedding to be pre-specified or utilise some type of heuristic to find a trade-off between dimensionality and embedding equality. To ensure a fair comparison, we test each method with different pre-specified values of embedding dimensionality ($d$) in the range $[1,10]$. For the two GP methods (GP-MaL and GP-MaL-LT), this corresponds to using between one and 10 trees per individual. We do not test larger dimensionalities as the performance of MaL methods generally converges by a dimensionality of 10 due to the reasonably low intrinsic dimensionality\footnote{The intrinsic dimension (ID) of a dataset is the number of dimensions required to accurately describe the important structure/characteristics of the data.} of most real-world data \cite{amsaleg2015estimating,facco2017estimating}.

\subsection{Datasets}
The performance of MaL methods is not only affected by the dimensionality of the data but also the number of instances and the number of intrinsic groupings (e.g.\ classes). Different domains and applications also have different feature distributions and interactions (e.g.\ in bioinformatics or text processing, sparse encodings are common). In light of this, we selected various datasets from a range of applications that contain different ratios of instances, features, and classes. These are listed in \cref{table:datasets}. \revisionOne{The feature range of all datasets was pre-scaled to $[0,1]$ to prevent any feature from having a greater impact on distance calculations (in the fitness function).}


		\renewcommand{\arraystretch}{1.1}
\begin{table}[t]
	\vspace{-1em}
	\caption{Classification datasets used for experiments.}
	\label{table:datasets}
	\centering
	\begin{tabularx}{.9\linewidth}{@{}Xrrrr@{}}
		\toprule
		Dataset & Instances & Features & Classes & Citation\\
		\midrule
		Image Segmentation & 2310 & 19 & 7 & \cite{uci}\\ 
		Ionosphere & 350 & 34 & 2 & \cite{uci}\\ 
		Dermatology & 358 & 34 & 6 & \cite{uci}\\ 
		Madelon & 2600 & 500 & 2 & \cite{guyon2004result}\\ 
		Isolet  & 1560 & 617 & 26 & \cite{uci}\\ 
		MFEAT & 2000 & 649 & 10 & \cite{uci}\\ 
		Yale & 165 & 1024 & 15 & \cite{yale2001}\\
		COIL20 & 1440 & 1024 & 20 & \cite{nene1996columbia}\\ 
		\revisionTwo{Bioresponse} & \revisionTwo{3751} & \revisionTwo{1776} & \revisionTwo{2} & \revisionTwo{\cite{bentzien2013crowd}}\\
		\revisionOne{Gisette} & 	\revisionOne{7000} & 	\revisionOne{5000} & 	\revisionOne{2} & 	\revisionOne{\cite{guyon2004result}}\\
		Tumour & 174 & 12533 & 11 & \cite{alanni2019tumor}\\ 
		Ovarian Cancer & 253 & 15153 & 2 & \cite{petricoin2002ovarian}\\ 
		\bottomrule
		
	\end{tabularx}
	\vspace{-1em}
\end{table}

\subsection{Parameter Settings}

	\begin{table}[t]
		\vspace{-1em}
	
	\centering
	\caption{GP methods' parameter settings.}
	\vspace{-.5em}
	\label{table:parameterSettings}
	\begin{tabularx}{0.95\linewidth}{ll X ll}
		
		\toprule
		Parameter& Setting && Parameter & Setting\\
		\cmidrule(r){1-2}  \cmidrule(l){4-5}
		Generations & 1000 && Population Size & 100\\
		Mutation & 20\% && Crossover & 80\% \\
		Min.\ Tree Depth & 2 && Max.\ Tree Depth & 14\\
		Elitism & top-10 && Pop. Initialisation & Half-and-half\\
		
		\bottomrule
	\end{tabularx}%
	\vspace{-1em}
\end{table}
To ensure our comparisons are as fair as possible, we use commonly used parameter settings for the baseline methods (as per the sci-kit-learn library \cite{scikit-learn}) and the same evolutionary parameters for the two GP approaches (shown in \cref{table:parameterSettings}).

GP-MaL-LT has an additional parameter: the number of neighbours to consider as a local neighbourhood ($K$). To test the sensitivity of $K$, we evaluated a range of possible $K$ values, sampled from the range $[10,100]$. Preliminary testing found that considering fewer than $10$ neighbours tended to reduce performance due to not giving a sufficient representation of the local topology. In contrast, more than $100$ became computationally much more expensive while also reducing performance due to the neighbourhood becoming too large. In our initial testing, we found that a $K$ of $30$ seemed to give an appropriate trade-off, and so we use this value when comparing it with the baseline methods. We provide further analysis of the sensitivity and guidance on the setting of $K$ in \cref{sensitivity}.

\section{Results and Discussion}
We present the results in three parts: firstly, by comparing the baseline GP-MaL method to the proposed GP-MaL-LT; then GP-MaL-LT compared against all baselines, and finally provide a sensitivity analysis of the effect of the $K$ parameter on the performance of GP-MaL-LT. For each set of results, we use the mean 10-fold classification accuracy as our evaluation criterion. \revisionOne{The full set of results are also provided as tables in the supplementary material. The supplementary material also shows the classification accuracy obtained when using $k$-nearest neighbour (instead of RF) as a classifier; $k$-NN shows a very similar pattern to RF, so we have omitted further analysis of those results in the main paper for brevity.}

\dsResults{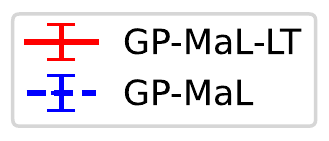}{Results for GP-MaL-LT and GP-MaL for the \revisionTwo{12} datasets for between one and 10 trees ($d=1$ to $d=10$). Each point on the plot shows the mean classification test accuracy across the 30 independent runs, with error bars representing the standard deviation. The maximum and minimum results of the 30 runs are shown by a  $\blacktriangle$ or a $\blacktriangledown$ respectively.}{.15}{.6}{.84}

\begin{table*}[tb!]
\centering
\revisionTwoPar{
\caption{p-values obtained from a Wilcoxon signed significance test (corrected using Hommel's method) testing for a difference in average classification accuracy between GP-MaL-LT and GP-MaL paired across the \revisionTwo{12} datasets for each tested value of $d$. GP-MaL-LT significantly outperformed GP-MaL for all $d$ at a significance level of $1\%$.}
\label{sigtestsGP}
\begin{tabular}{lrrrrrrrrrr}
\toprule
\multicolumn{1}{r}{$d= $} &        1  &        2  &        3  &        4  &        5  &        6  &        7  &        8  &        9  &        10 \\
\midrule
Raw p-values       &  0.001465 &  0.000488 &  0.001465 &  0.000488 &  0.000488 &  0.000488 &  0.000488 &  0.002441 &  0.001465 &  0.000977 \\
Corrected p-values &  0.002441 &  0.001953 &  0.002441 &  0.001953 &  0.001953 &  0.001953 &  0.001953 &  0.002441 &  0.002441 &  0.002197 \\
\bottomrule
\end{tabular}
}
\vspace{-1em}
\end{table*}

\begin{table*}[tb]
\vspace{-1em}
\centering
\revisionTwoPar{
\caption{p-values obtained from a Wilcoxon signed significance test (corrected using Hommel's method) testing for a difference in average run time between GP-MaL and GP-MaL-LT paired across dimensionalities for each dataset. No significant difference was found at a significance level of $10\%$.}
\label{sigtestsGPtime}
\begin{tabularx}{\textwidth}{lXXXXXXXXXXXX}
\toprule
{} &  Image Seg.\ &  Iono.\ &  Derm.\ &  Madelon &   Isolet &   MFEAT &     Yale &   COIL20 &  Tumour &  Ovarian &  Bio.\ &  Gisette \\
\midrule
Raw p-values       &             0.03125 &     0.03125 &      0.03125 &  0.03125 &  0.03125 &  0.0625 &  0.03125 &  0.03125 &     1.00000 &         0.84375 &      0.03125 &  0.03125 \\
Corrected p-values &             0.12500 &     0.12500 &      0.12500 &  0.12500 &  0.12500 &  0.1875 &  0.12500 &  0.12500 &     1.00000 &         1.00000 &      0.12500 &  0.12500 \\
\bottomrule
\end{tabularx}
}
\vspace{1em}
\end{table*}

\subsection{GP-MaL-LT compared to GP-MaL}

\Cref{fig:gps} shows the results of the two GP methods on each of the \revisionTwo{12} classification datasets. The datasets are arranged from left-to-right and top-to-bottom in order of increasing dimensionality. 

On the majority of the datasets, GP-MaL-LT shows a clear and significant improvement over GP-MaL, with the two methods' error bars often not overlapping, especially at the lowest dimensionalities ($d < 3$). On several of the datasets (Image Segmentation, Ionosphere, COIL20), the performance of the two GP methods convergence (at high accuracy) for $d > 4$, indicating that the data has a low intrinsic dimensionality. 

For the Isolet, Tumour, \revisionTwo{Bioresponse,} and MFEAT datasets, GP-MaL-LT consistently outperforms GP-MaL for nearly all values of $d$, with a particularly large performance difference shown at low dimensionality on MFEAT and Isolet. On Isolet, GP-MaL-LT achieves nearly double the accuracy of GP-MaL at $d=1$, which is a substantial improvement for such a small embedding. GP-MaL-LT is also much more consistent than GP-MaL, with lower standard deviations and much higher minimum results across the board. These results clearly demonstrate the value of the proposed approach in improving the quality and consistency of manifold learning. 

On the Yale, Ovarian Cancer, and Dermatology datasets, the two GP methods do not provide significantly different performance. However, GP-MaL-LT tends to be more consistent, with much higher minimum accuracies on Ovarian Cancer and Dermatology at low dimensionalities. The Yale dataset has the fewest number of instances of all datasets (165) --- this may be a factor in the similar performance of the two methods; focusing on optimising the preservation of local topology is less crucial when the dataset is small (i.e.\ reasonably ``local" to start with). Very high-dimensional datasets such as Ovarian Cancer (15,153 features) are particularly challenging for GP methods due to the substantial number of possible combinations of features that can be included in a GP tree \cite{tran2019genetic}, with several authors choosing to perform feature selection methods to reduce the search space of GP \cite{chen2017feature,mei2017efficient}.

Madelon is an artificial dataset designed for testing feature selection methods, which also has beneficial properties for evaluating MaL algorithms. Of the 500 features, only five are meaningful to the class labels; 15 features are redundant (linear combinations of meaningful features), and the remaining 480 are noise. A good MaL algorithm should recognise that only five features are useful for characterising this dataset. This is the pattern that we see with GP-MaL-LT: the performance steadily increases until an embedded dimensionality of five (i.e.\ on average, all five features are ``selected"), at which point it plateaus as there is no further information in the remaining features. GP-MaL also follows a similar trend but has a much wider variation in performance than GP-MaL-LT, along with lower mean accuracy for more than $d \geq 3$ dimensions. 
 
\subsubsection*{Statistical significance testing} \revisionOnePar{To further compare the two GP methods, we performed a Wilcoxon signed-rank significance test for each dimensionality ($d$) across the \revisionTwo{twelve} datasets, with Hommel's method used to correct for type-1 errors. The p-values produced by this test are shown in \cref{sigtestsGP}. These tests showed that GP-MaL-LT significantly outperforms GP-MaL at every setting of $d$ at $\alpha=0.01$ (with 99\% confidence). \Cref{fig:gps} also shows that GP-MaL-LT has lower variance in performance compared to GP-MaL, which makes it a more reliable MaL algorithm to use in practice where only one GP run may be performed. 
}

\subsubsection*{Runtime analysis} \revisionOnePar{We also empirically compared the mean run times of GP-MaL-LT compared to GP-MaL across the datasets. We tested for a difference in run time between the two methods using the same significance testing procedure as above; the computed p-values are shown in \cref{sigtestsGPtime}. The testing shows that there is no significant difference in run time between GP-MaL and GP-MaL-LT at a $10\%$ significance level (\cref{sigtestsGPtime}), despite the large improvement in classification performance of GP-MaL-LT compared to GP-MaL. The full run time results are included in the supplementary material.}

\dsResults{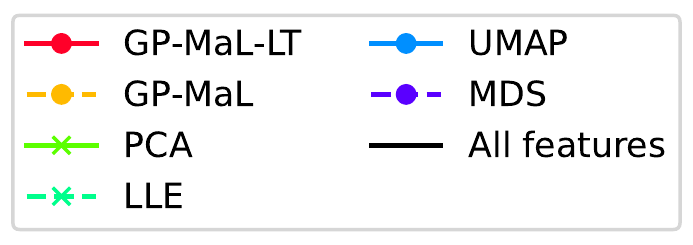}{Results for GP-MaL-LT and all baseline methods for the \revisionTwo{12} datasets for between one and 10 dimensions/trees ($d$). Each point on the plot shows the mean classification test accuracy across the 30 independent runs (for the stochastic methods). The black horizontal line represents the accuracy attained when using all features of the dataset, providing an approximate upper bound on performance.}{.3}{1}{.69}

\subsection{GP-MaL-LT compared to all baselines}
The second set of plots (\cref{fig:compared}) shows the performance of GP-MaL-LT compared to all baselines across the \revisionTwo{12} datasets. We also include the accuracy attained when using all features of the dataset, represented by a black horizontal line --- this provides an upper bound on the manifold learning performance, assuming that all features are relevant to the class labels. To maintain readability, we do not include error bars or maximum and minimum results. 

Except for the Isolet, Tumour, Gisette, and Ovarian Cancer datasets, GP-MaL-LT is clearly competitive with all baselines for $d > 8$ and often outperforms several baseline methods at lower dimensionalities. On Image Segmentation, Ionosphere, Dermatology, Madelon, COIL20, \revisionTwo{and Bioresponse,} it is first-equal or second at $d=1$ while being no worse than third-best on the remaining datasets. On Madelon, it is the only method aside from PCA which can successfully ``solve" the problem by selecting the five relevant features. PCA performs very well due to its selection of successive principal components that maximise the residual variance, but the other non-GP MaL methods struggle due to their use of an explicit mapping that is much more prone to treating noise in the data as part of the manifold structure. 

On the very high-dimensional Tumour, Gisette, and Ovarian Cancer datasets, GP-MaL-LT struggles somewhat but outperforms MDS and PCA at a dimensionality of one or two, despite the large number of possible features to choose from. We are confident that incorporating feature selection approaches into future work will significantly improve the performance of GP-based MaL on these very high-dimensional datasets --- the maximum accuracies achieved by GP-MaL-LT are promising. On Isolet, GP-MaL-LT is significantly outperformed by UMAP and by LLE when the dimensionality is below five. It is, however, superior to the remaining MaL baselines at low dimensionalities and similar at higher dimensionalities. Isolet is a challenging dataset for MaL algorithms as it contains 26 distinct classes (letters of the alphabet), each of which has 60 instances.

While the performance of the MaL methods is dataset-dependent, there are still clear patterns. Given that GP-MaL-LT is an implicit mapping method, it has very promising performance compared to the baseline methods, which all use an explicit (non-parametric) mapping. While the state-of-the-art UMAP method did outperform the proposed method on high-dimensional datasets, GP-MaL-LT can match or exceed the performance of other baselines. UMAP results from over 20 years of research into explicit mapping methods; research on implicit mapping MaL methods is in its infancy in comparison. GP-MaL-LT has high maximum accuracy (the $\blacktriangle$s in \cref{fig:gps}), indicating it has the potential to match (or even exceed) UMAP's performance if its variance can be reduced.

\dsResults{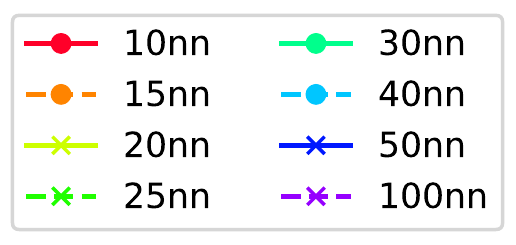}{Effect of the \textit{K} parameter in GP-MaL-LT on attained test accuracy. Each line represents a different setting of $K$, from $K=10$ (10 nearest-neighbours) to $K=100$ (100nn). As before, this is shown for $d$ in the range $[1,10]$ to show patterns across different sizes of embedding.}{.23}{1}{.76}

Also noteworthy is the performance of LLE compared to GP-MaL-LT, given they optimise the preservation of local topology. Across the \revisionTwo{12} datasets, GP-MaL-LT is better than LLE on five (Image Segmentation, Ionosphere, Dermatology, Madelon, and Yale), while LLE outperformed GP-MaL-LT on three (MFEAT, Tumour, and Ovarian Cancer). Generally speaking, LLE is superior on high-dimensional datasets. This further reinforces our theory that the large feature space is much easier for explicit mapping methods (e.g.\ LLE) to cope with: when the MaL method does not need to find a function that maps the input to the embedded space, it is not nearly as affected by an increase in intrinsic dimensionality. In future work, we will seek to improve GP-MaL-LT by addressing this discrepancy by using feature selection/weighting techniques to make the feature space more constrained.

\subsubsection*{Statistical significance testing} \revisionTwoPar{We also tested for a statistically significant difference in classification accuracy between GP-MaL-LT and the baseline methods for each dimensionality ($d$) across the twelve datasets. For each dimensionality, a Friedman test was performed to test for any difference between the methods. Where there was a significant difference (a p-value of less than $\alpha=0.05$), we then performed posthoc analysis using the Holm-Bonferroni method with GP-MaL-LT as the control method. The Friedman test showed a significant difference for all dimensionalities except for $d=7$ (which had a p-value of $0.068$). The posthoc analysis showed a significant difference (at $\alpha=0.05$) between GP-MaL-LT and PCA, and between GP-MaL-LT and MDS, for $d=1$ and $d=2$. This analysis is consistent with the graphed results, which consistently show PCA and MDS performing very poorly at low values of $d$. There were no other statistically significant findings in the posthoc analysis; the differences detected by the Friedman test for higher values of $d$ is likely between pairs of baseline methods, which we do not test for, as our focus is on comparisons with our proposed method (GP-MaL-LT). The full results of this testing are included in the supplementary material.}

\subsection{Sensitivity analysis of the \textit{K} parameter in GP-MaL-LT}
\label{sensitivity}
The mean test accuracy attained by GP-MaL-LT for different settings of the $K$ parameter is shown in \cref{fig:nns}. In most cases, GP-MaL-LT is insensitive to the value of $K$, with a few notable exceptions. 

There is a decrease in performance when using $K=100$ on the Dermatology dataset, but relatively little difference between other values of $K$. The Dermatology dataset has several classes containing few instances: for example, the \textit{pityriasis rubra pilaris} class contains only 20 instances. When $K=100$, an instance's neighbourhood will likely contain many different classes, which may cause GP-MaL-LT to optimise the ordering of the closest neighbours less strongly.

A high $K$ value is clearly beneficial for the Madelon dataset. A low value of $K$ makes GP-MaL-LT more prone to the effects of noise. With a small neighbourhood, the effect of noisy features is more likely to warp the topology, whereas large neighbourhoods prevent this by considering a larger number of neighbours.

On the Yale and Tumour datasets, there is a noticeable decrease in accuracy as $K$ is increased, particularly to $K=100$. This is most evident on Tumour at $d=1$, with performance nearly halving from $k=10$ to $k=100$. This is perhaps not unexpected, as these are the two smallest datasets tested --- increasing $K$ to $100$ means that an instance's neighbourhood will comprise around half the dataset, diminishing the advantages of the proposed approach.  

\begin{figure}[t!]
	\vspace{-1em}
	\centering
	\includegraphics[width=.9\linewidth]{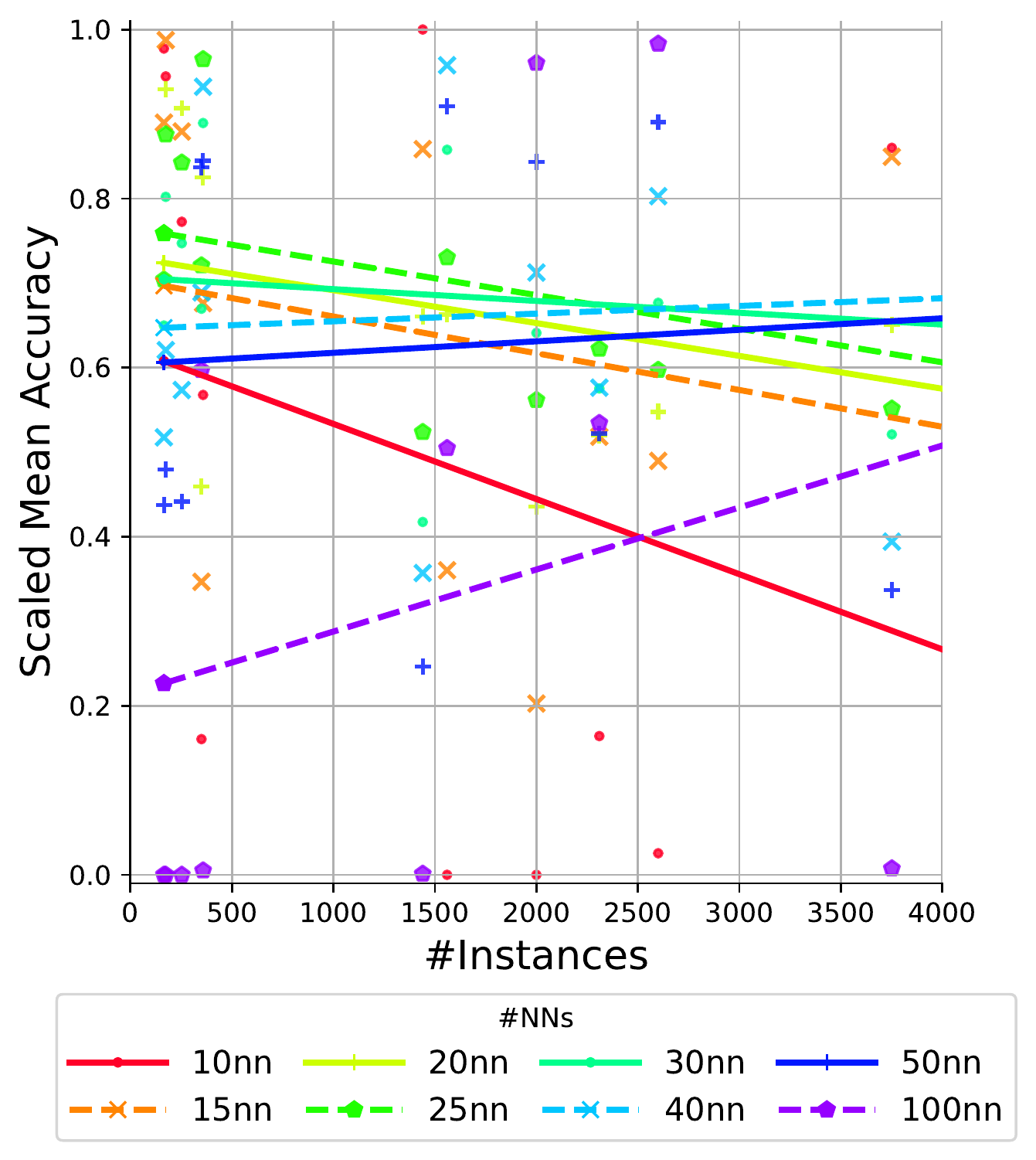}\vspace{-.5em}
    \caption{The effect of the $K$ parameter in GP-MaL-LT on classification test accuracy as dataset size increases (x-axis limited to 4000 for readability).}
    \label{kTests}
	\vspace{-1em}
\end{figure}

Indeed, there appears to be a positive correlation between the number of instances in a dataset and the best value of $K$. To analyse this further, \cref{kTests} shows the scaled accuracy attained by each tested $K$ value as a function of dataset size. For each dataset, the mean accuracy was computed for each value of $K$ (across the 10 different settings of the number of trees used). Then these were scaled between $0$ and $1$ to enable a clear comparison between different datasets. In other words, a scaled mean accuracy of $0$ indicates that value of $K$ gave the worst mean accuracy for that dataset, whereas $1$ indicates it was the best. A line-of-best-fit is also shown for each value of $K$ to demonstrate the overall tendency. 

\Cref{kTests} confirms our suspicions: on a small number of instances, $K=100$ performs very poorly, but on large datasets, it is often the best value. At the other extreme, $K=10$ is reasonably close to the best performing $K$ values on small datasets but is clearly too small for large datasets. Indeed, this pattern is true across all tested values of $K$, with smaller values being better on smaller datasets and larger values on larger data. For datasets smaller than $\approx 1500$ instances, $K=25$ gives the best results; datasets with 1500--2300 instances perform the best with $K=50$ or $K=40$; and on larger datasets, $K=100$ is the most appropriate. Intermediary values of $K$ such as $K=30$ give the best ``overall" results --- $K=30$ is never worse than more than three of the other seven settings and never has a scaled mean accuracy below $0.65$. If only a single $K$ value was chosen for all datasets, $K=30$ is likely the best choice. Another approach may be to choose $K$ as a fraction of the dataset size so that neighbourhoods are always the same proportion of the dataset.

\section{Further Analysis}
To better understand the difference in performance between GP-MaL-LT and LLE (the two local cost approaches), we provide visualisations of the results of these methods on two datasets. The first, COIL20, is an example where GP-MaL-LT is clearly superior at a single dimension but is slightly outperformed at higher dimensionalities. On the second, MFEAT, LLE outperforms across all tested dimensionalities. For each dataset, we focus our visualisation on the one- and two-dimensional cases: this is where the biggest difference in performance is seen, and the results at these dimensionalities can also be visualised directly without the use of a visualisation technique. In all our visualisations, we exclude outliers (defined as instances with feature values in the top or bottom 3\% OF values) to focus on the main patterns in the data. For the two-dimensional visualisations, we add a small amount of random jitter to the data so that multiple instances with the same coordinates can be seen. 

\onecf{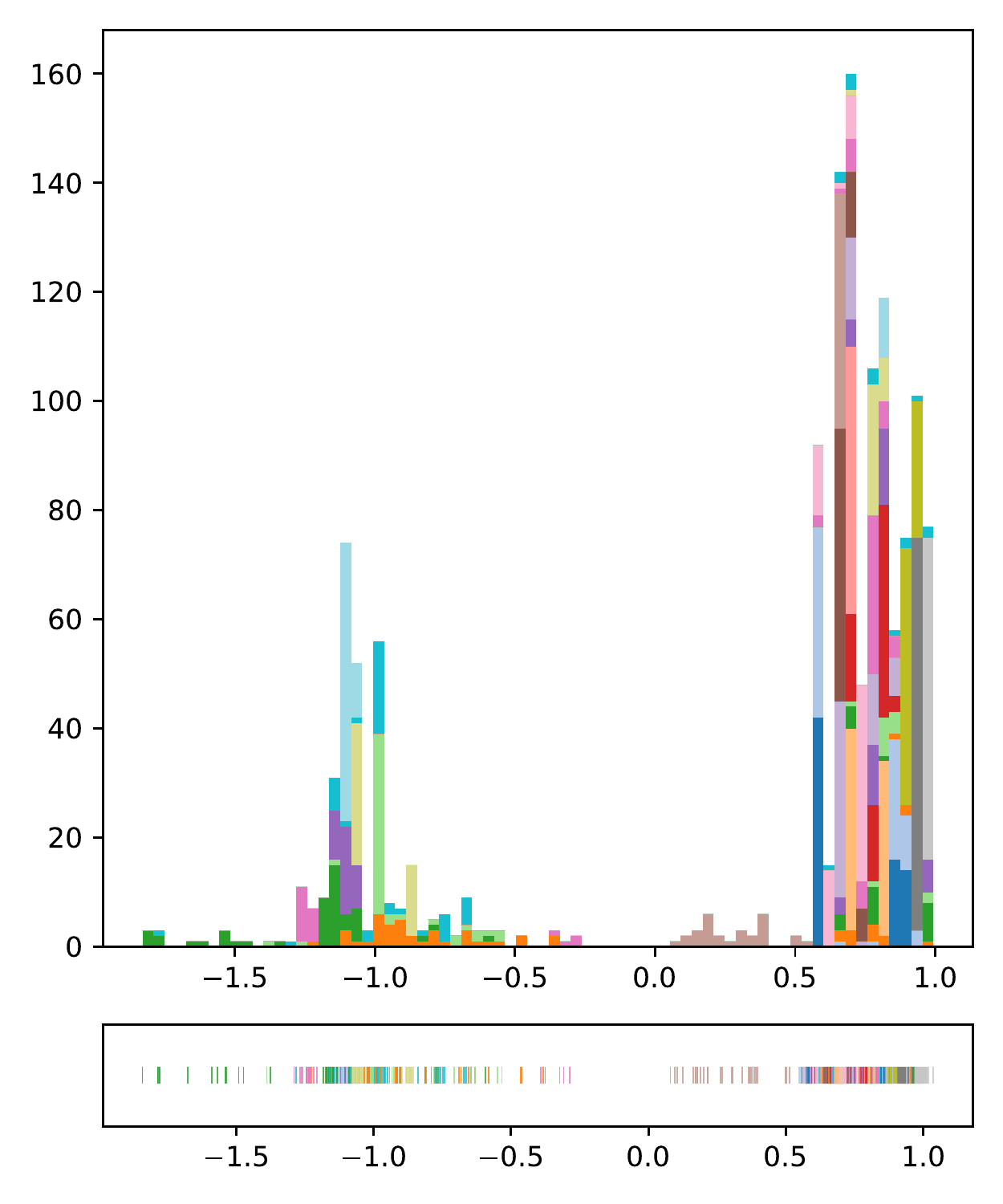}{COIL20}
\twocf{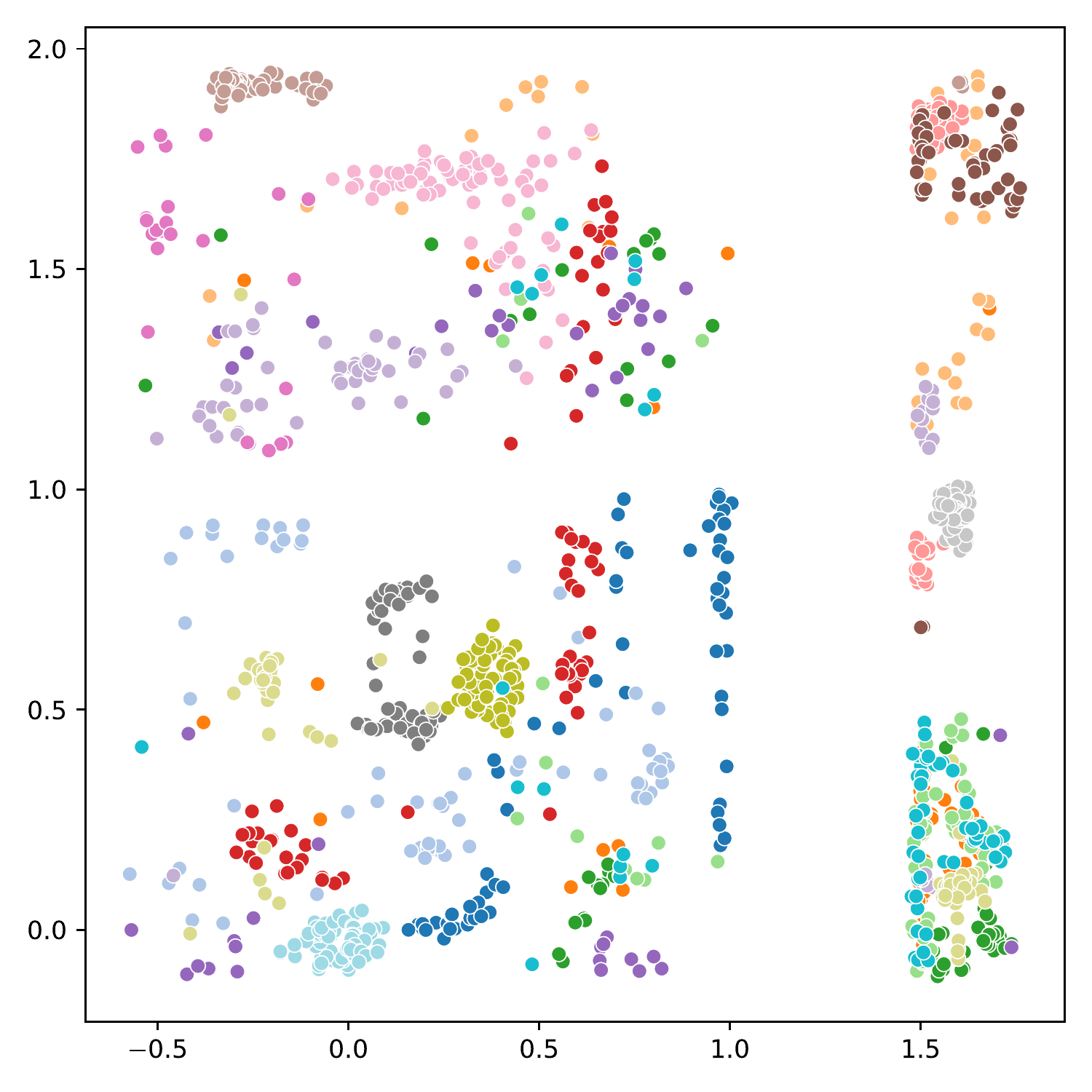}{COIL20}

\subsection{COIL20}

The one-dimensional visualisations for the COIL20 dataset are shown in \cref{fig:COIL20-1d-results}. We show the median and best GP-MaL-LT performance to demonstrate the typical and potential performance of the proposed method. For each result, a histogram of the feature values, as well as a number line, are provided, where both are coloured according to the class labels of the dataset. At one embedded dimension, LLE clearly struggles, separating one class very well but failing to separate the remaining classes at all. In contrast, GP-MaL-LT can separate the different classes into different groups, with the brown class on the median result and the dark blue, pink, yellow-green, and light blue classes on the best result being nearly completely separated from other classes. 

At two embedded dimensions (\cref{fig:COIL20-2d-results}), LLE separates the classes much more effectively, with slightly superior classification accuracy to the best result of GP-MaL-LT. Two classes are particularly well separated, with the remaining classes appearing very close together in the embedded space along the x-axis. In contrast, the GP-MaL-LT methods give a much wider spread of instances belonging to the same class while also distributing different classes more widely around the embedded space. LLE uses a much more constrained neighbourhood than GP-MaL-LT, which means it can group instances of the same class very tightly together, at the expense of losing some of the overall interclass topology that GP-MaL-LT retains. This demonstrates the intrinsic trade-off in MaL approaches: prioritising the preservation of the local structure retains (sub-)classes in the data at the expense of the overall interclass (global) topology.   

\onecf{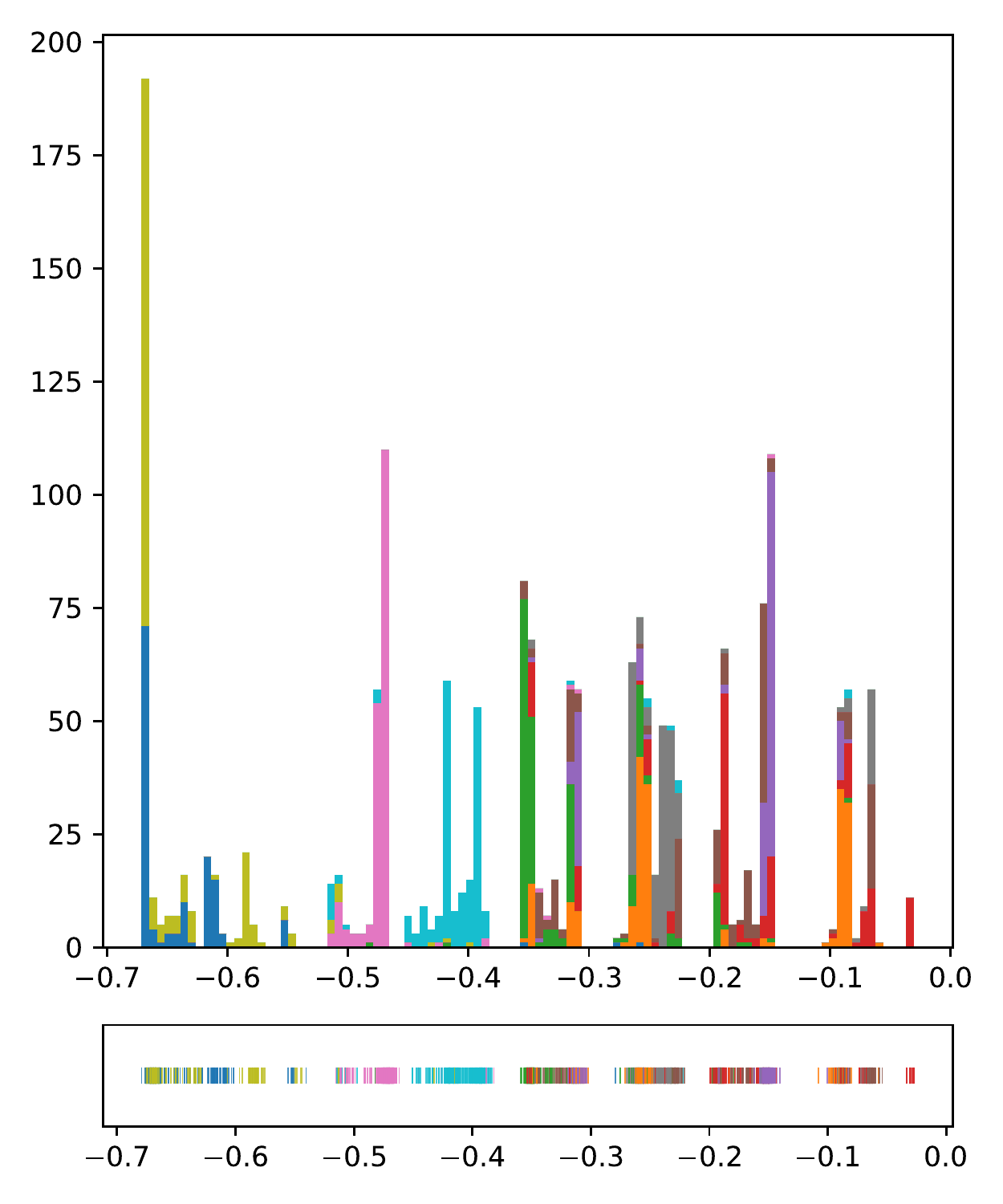}{MFEAT}
\twocf{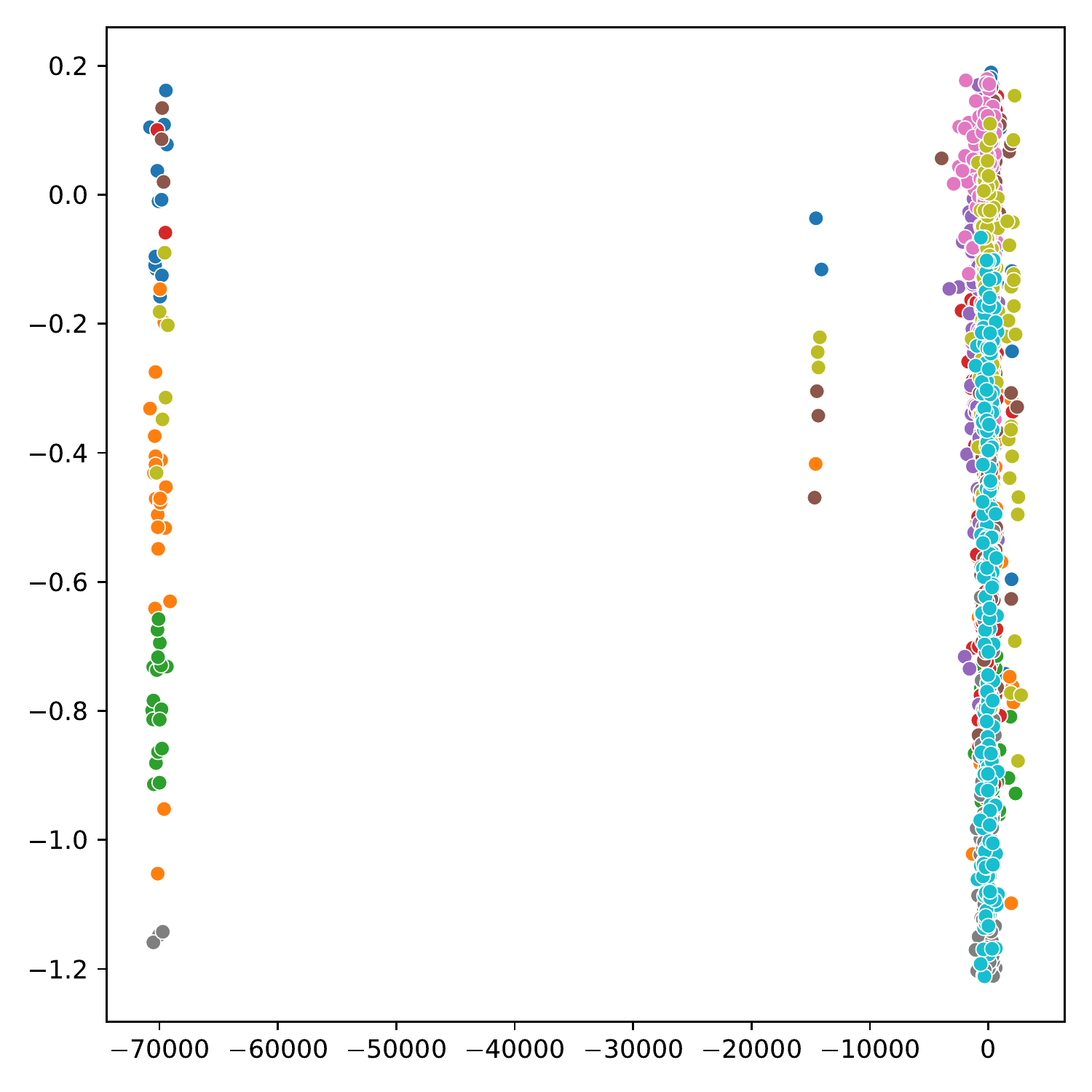}{MFEAT}

\subsection{MFEAT}
Using one embedded dimension on the MFEAT dataset (\cref{fig:mfatCombined-1d-results}), LLE can separate a number of the classes very clearly, with the remaining classes overlapping with others. GP-MaL-LT does not give such clear separations but appears to group common classes reasonably well. At two dimensions (\cref{fig:mfatCombined-2d-results}), LLE produces a similar topology to in one dimension but separates the purple class more clearly. The best result of GP-MaL-LT (which achieves $\approx 90\%$ accuracy) is also able to group instances from the same class together well, albeit with more overlap between closely related classes. As with the COIL20 dataset, GP-MaL-LT better preserves the relationships within a single class than LLE --- most notably, the green class is spread along $y=-13$ to $y=-11$ in the best GP-MaL-LT result, but is nearly entirely obscured by the overlap with other classes in LLE. The median result for GP-MaL-LT is less clear, but the same general patterns are evident. Further research will aim to improve the consistency of GP-MaL-LT so that results similar to the best result on MFEAT can be obtained more reliably.

	\begin{figure}[t!]
		\vspace{-1em}
		\centering
		\includegraphics[width=.7\linewidth]{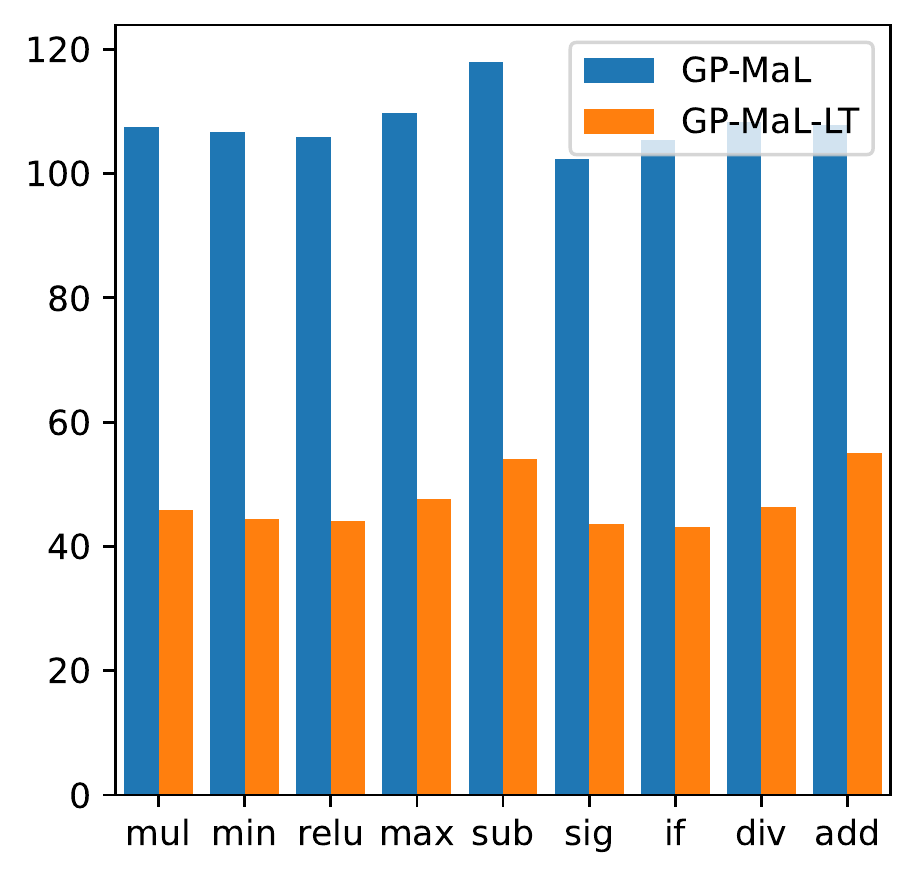}
		\vspace{-1em}
		\caption{\revisionOne{Average number of each function used for each GP individual for both GP-MaL and GP-MaL-LT. GP-MaL-LT uses many fewer functions and so produces much smaller trees.}}
				\label{gpVisComp1}
		\vspace{-2em}
	\end{figure}
	
\revisionOnePar{
\subsection{Interpretability of GP-MaL-LT}
A major benefit of implicit mapping methods is their potential for interpretability: their functional mapping can be directly examined to understand how they operate. However, if these mappings are overly large and complex, they are still difficult to understand. To evaluate the potential interpretability of the two GP methods, we compared the average number of terminals and functions used by an individual produced by each of GP-MaL and GP-MaL-LT. As shown in \cref{gpVisComp1,gpVisComp2}, GP-MaL-LT uses fewer than 40\% as many feature terminals as GP-MaL, and between 40\% and 50\% the number of function nodes. This, in conjunction with the better classification performance, suggests that GP-MaL-LT is much more effective than GP-MaL as a MaL method on the tested problems.

    \begin{figure}[t]
		\vspace{-1em}
		\centering
		\includegraphics[width=.75\linewidth]{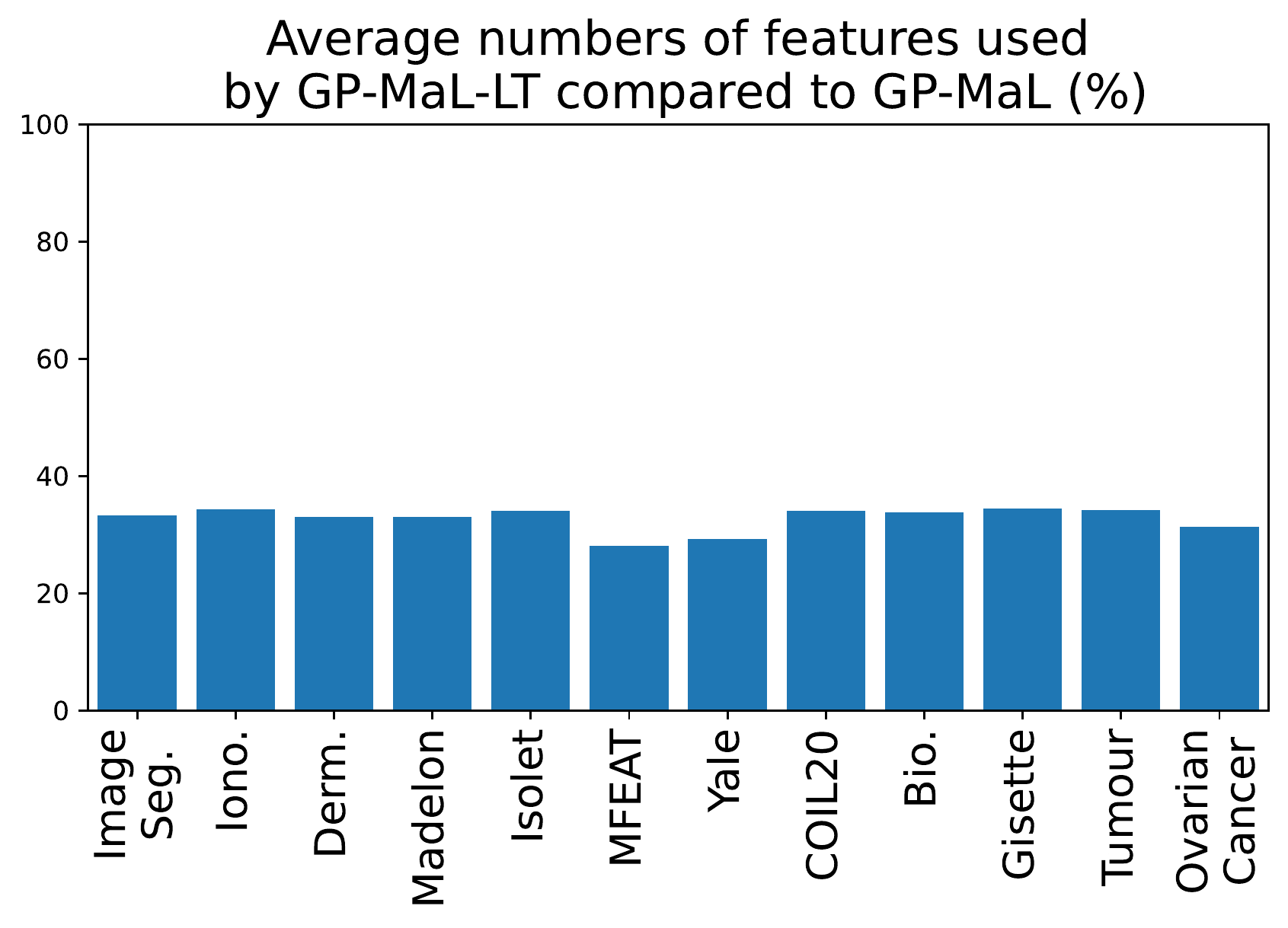}
		\vspace{-1em}
		\caption{\revisionOne{Comparison between the mean number of features used by GP-MaL-LT and GP-MaL. GP-MaL-LT consistently uses fewer than 40\% the feature terminals of GP-MaL.}}
		\vspace{-1em}
		\vspace{-1em}
		\label{gpVisComp2}
	\end{figure}
}

\begin{figure*}[tbh!]
\vspace{-0.5em}
\centering

\begin{align*}
{PC}_1 = -0.121 \times \text{f493}-0.116 \times \text{f508}-0.115 \times \text{f494}-0.115 \times \text{f478}-0.114 \times \text{f479}-0.111 \times \text{f548}-0.109 \times \text{f507} ...\\
{PC}_2 = -0.133 \times \text{f403}-0.128 \times \text{f418}+0.121 \times \text{f408}-0.121 \times \text{f417}-0.120 \times \text{f388}+0.119 \times \text{f423}-0.118 \times \text{f404} ...\\
{PC}_3 = +0.130 \times \text{f488}-0.127 \times \text{f540}-0.123 \times \text{f525}+0.123 \times \text{f503}+0.123 \times \text{f489}-0.123 \times \text{f539}-0.119 \times \text{f555} ...\\
{PC}_4 = -0.128 \times \text{f553}-0.123 \times \text{f569}+0.121 \times \text{f418}+0.121 \times \text{f433}-0.119 \times \text{f568}-0.119 \times \text{f554}+0.115 \times \text{f403} ...\\
{PC}_5 = +0.159 \times \text{f413}+0.158 \times \text{f414}+0.157 \times \text{f429}+0.152 \times \text{f428}-0.138 \times \text{f373}-0.137 \times \text{f372}-0.132 \times \text{f359} ...\\
{PC}_6 = -0.191 \times \text{f578}-0.186 \times \text{f592}-0.172 \times \text{f591}-0.170 \times \text{f577}-0.166 \times \text{f593}-0.164 \times \text{f579}-0.136 \times \text{f411} ...\\
{PC}_7 = -0.176 \times \text{f468}-0.170 \times \text{f469}-0.160 \times \text{f467}-0.152 \times \text{f484}-0.148 \times \text{f470}-0.146 \times \text{f483}-0.140 \times \text{f485} ...
\end{align*}
\vspace{-1.5em}
\revisionOne{\caption{The first seven principal components resulting from PCA performed on the MFEAT dataset (Random Forest accuracy: 94.65\%). Note that only the seven largest weights (out of 649) are shown for each component.}}
\label{pcaMFEAT}
%
\vspace{2em}
\newcommand{\boldm}[1] {\mathversion{bold}#1\mathversion{normal}}

\begin{minipage}[b][5.12in][t]{0.4\textwidth}%
     \begin{overpic}[scale=0.35]{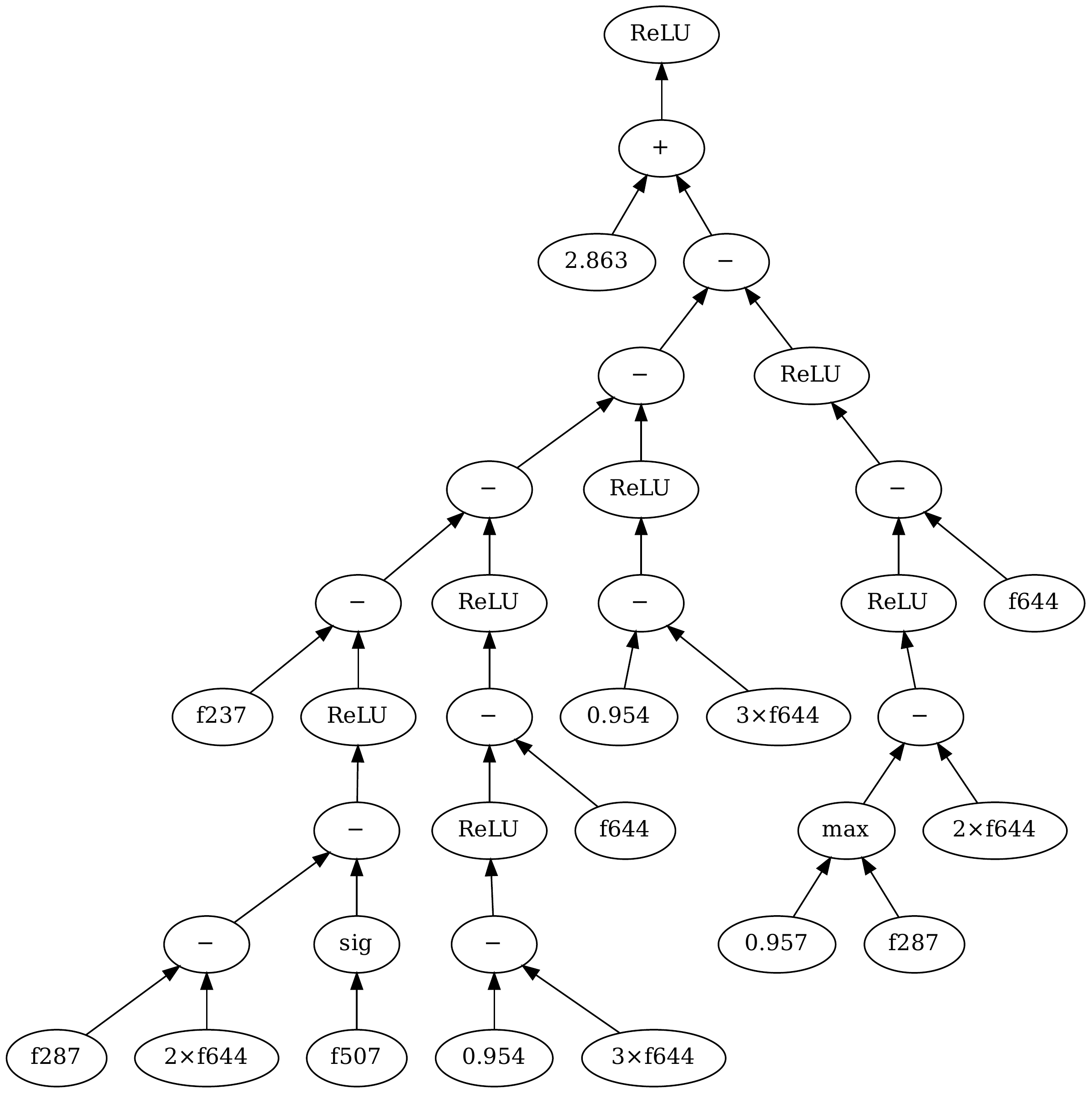}
     \put(5,94){\includegraphics[scale=0.35]{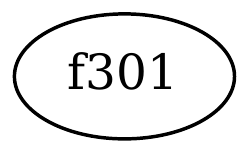}}  
     \put(30,94){\includegraphics[scale=0.35]{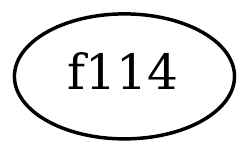}}
      \put(85,94){\includegraphics[scale=0.35]{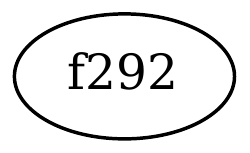}}
     \put(110,94){\includegraphics[scale=0.35]{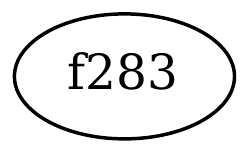}} 
     \put(8, 102){\boldm $T_1$}
     \put(33,102){\boldm $T_2$}
     \put(58, 102){\boldm $T_3$}
     \put(88, 102){\boldm $T_4$}
     \put(113,102){\boldm $T_5$}
  \end{overpic}
\end{minipage}
  \hfill
  \begin{overpic}[scale=0.35]{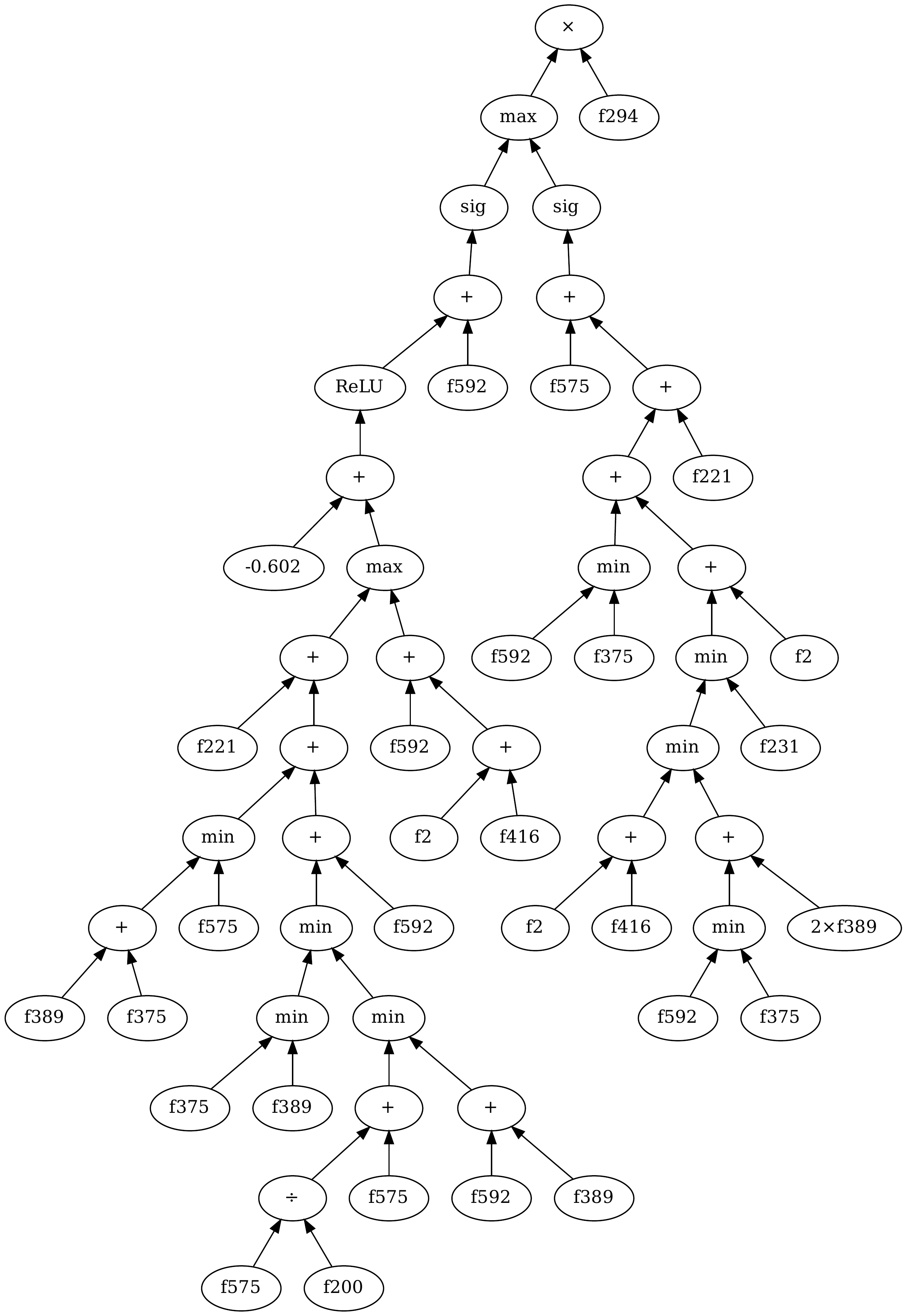}
   \put(20,96){\includegraphics[scale=0.35]{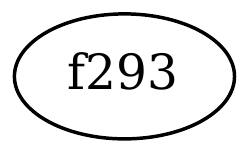}}
   \put(22,101.5){\boldm $T_6$}
   \put(42,101.5){\boldm $T_7$}
  \end{overpic}


    \revisionOne{\caption{The seven GP trees found by a run of GP-MaL-LT on the MFEAT dataset (Random Forest accuracy: 94.04\%).}}
    \vspace{-1em}
    \label{gpMFEAT}
\end{figure*}

PCA is also often regarded as a \textit{grey box} algorithm, as the principal components that it produces can potentially be interpreted by examining the magnitude of their weights for each feature. To better understand how the interpretability of GP-MaL-LT compares to PCA, we have provided an example model produced by each that has similar accuracy for $d=7$ on the MFEAT dataset \revisionTwo{ in \cref{pcaMFEAT,gpMFEAT}}. \revisionTwoPar{The trees produced by GP-MaL-LT were simplified through the use of algebraic simplification algorithms and manual inspection.  Five of the evolved GP trees are extremely simple, using only a single feature each. The remaining two trees are more complex, but the whole  individual uses only 19 unique features (with 57 occurrences) and 69 functions (before simplification) --- in contrast, a single component produced by PCA has 649 weights (one per feature). The use of parsimony pressure and other bloat control methods was outside the scope of this paper, but we expect that using these in future work would make even smaller GP models without any loss in accuracy.

It is somewhat counter-intuitive that five of the trees in \cref{gpMFEAT} are single features --- we would generally expect more complex structures to be needed to preserve sufficient structure. We investigated this phenomenon further and found that two of the trees ($T_4$ and $T_6$) represent the first two Karhunen–Lo\`{e}ve coefficients of the original image data. The Karhunen–Lo\`{e}ve transformation is very closely related to PCA in that it decomposes the image data into orthogonal components. Hence, the first two Karhunen–Lo\`{e}ve coefficients represent the two highest entropy components and thus have the highest information of the decomposed components. Given that manifold learning seeks to preserve as much information as possible, it is encouraging that GP-MaL-LT has selected these two features in this example. 

To further understand the contribution of each of the trees in \cref{gpMFEAT} in preserving local topology, we investigated the change in fitness as each tree is added to the embedding. $T_3$ has the lowest fitness (best preservation of local topology) of any of the trees, at $\approx 13.23$ (recall that for $K=30$ neighbours, fitness is in $[0,3]$). Adding $T_4$ improves the fitness to $11.07$; adding $T_7$ then improves it to $9.15$; adding $T_6$ then improves it to $7.53$; adding $T_2$ improves it to $6.41$; adding $T_5$ improves it to $5.54$; and finally, adding the seventh tree, $T_1$, improves the fitness to $4.98$. This shows that each of the seven trees contributes to maximally preserving local topology and that GP-MaL-LT can automatically discover a set of complementary trees for a given dimensionality.
}

\section{Conclusions}

This paper introduced a new approach to performing GP-based MaL that focuses on preserving local topology to improve the quality of GP-based MaL in applications where local topology preservation is essential (such as classification or segmentation tasks). A novel fitness function was derived based on the ordering of instances in a local neighbourhood, and an approximation technique was introduced to make the proposed GP-MaL-LT algorithm computationally feasible. The proposed method was compared to several MaL baseline and state-of-the-art approaches across a representative set of classification problems to evaluate how well the local topology is preserved. GP-MaL-LT showed a clear improvement over GP-MaL across all the tested datasets, obtaining higher test accuracy (particularly at low embedding dimensionality) and lower variance in performance. GP-MaL-LT is competitive and often better than the well-known LLE method, which also focuses on preserving the local topology. This was despite GP-MaL-LT being an \textit{implicit} mapping method, which produces a functional (parameterised) mapping between the original and embedded spaces. Finding such a mapping is inherently more difficult compared to freely optimising the embedded space (as the non-GP baseline methods do) but has clear advantages in terms of model reusability and interpretability.

Given that GP-based MaL is an emerging topic, there are many promising areas of future research. In this paper, we demonstrated the clear potential of the proposed approach to compete with state-of-the-art methods. Still, we found that the variance in results needed to be decreased on very high-dimensional data. In future work, we will investigate the use of feature selection or feature weighting to decrease or optimise the feature space that GP searches in. We would also like to investigate an evolutionary multi-objective approach that balances the two conflicting objectives of preserving local and global topology to better understand the trade-offs between them. \revisionOne{Another promising direction is developing new fitness functions that are \textit{differentiable}, as this would then allow for efficient and effective gradient-based optimisation of coefficients in the model}. Finally, given that GP-based MaL is an implicit mapping approach, we would like to \revisionOne{further} investigate the interpretability of the learned models, with a focus on decreasing their complexity to enable human understanding of the intrinsic relationships in data.


\bibliographystyle{IEEEtran}
\bibliography{biblo}

\begin{thebibliography}{10}
\providecommand{\url}[1]{#1}
\csname url@samestyle\endcsname
\providecommand{\newblock}{\relax}
\providecommand{\bibinfo}[2]{#2}
\providecommand{\BIBentrySTDinterwordspacing}{\spaceskip=0pt\relax}
\providecommand{\BIBentryALTinterwordstretchfactor}{4}
\providecommand{\BIBentryALTinterwordspacing}{\spaceskip=\fontdimen2\font plus
\BIBentryALTinterwordstretchfactor\fontdimen3\font minus
  \fontdimen4\font\relax}
\providecommand{\BIBforeignlanguage}[2]{{%
\expandafter\ifx\csname l@#1\endcsname\relax
\typeout{** WARNING: IEEEtran.bst: No hyphenation pattern has been}%
\typeout{** loaded for the language `#1'. Using the pattern for}%
\typeout{** the default language instead.}%
\else
\language=\csname l@#1\endcsname
\fi
#2}}
\providecommand{\BIBdecl}{\relax}
\BIBdecl

\bibitem{lee2007nonlinear}
J.~A. Lee and M.~Verleysen, \emph{Nonlinear dimensionality reduction}.\hskip
  1em plus 0.5em minus 0.4em\relax Springer Science \& Business Media, 2007.

\bibitem{bengio2013representation}
Y.~Bengio, A.~C. Courville, and P.~Vincent, ``Representation learning: {A}
  review and new perspectives,'' \emph{{IEEE} Trans. Pattern Anal. Mach.
  Intell.}, vol.~35, no.~8, pp. 1798--1828, 2013.

\bibitem{espadoto2019towards}
M.~{Espadoto}, R.~M. {Martins}, A.~{Kerren}, N.~S.~T. {Hirata}, and A.~C.
  {Telea}, ``Towards a quantitative survey of dimension reduction techniques,''
  \emph{IEEE Transactions on Visualization and Computer Graphics}, vol.~27,
  no.~3, pp. 2153--2173, 2021.

\bibitem{maatenTSNE}
L.~van~der Maaten and G.~E. Hinton, ``Visualizing high-dimensional data using
  t-{SNE},'' \emph{Journal of Machine Learning Research}, vol.~9, pp.
  2579--2605, 2008.

\bibitem{2018arXivUMAP}
\BIBentryALTinterwordspacing
L.~{McInnes}, J.~{Healy}, and J.~{Melville}, ``{UMAP:} uniform manifold
  approximation and projection for dimension reduction,'' \emph{ArXiv
  e-prints}, p. arXiv:1802.03426, 2020. [Online]. Available:
  \url{http://arxiv.org/abs/1802.03426}
\BIBentrySTDinterwordspacing

\bibitem{roweis2000lle}
S.~T. Roweis and L.~K. Saul, ``Nonlinear dimensionality reduction by locally
  linear embedding,'' \emph{Science}, vol. 290, no. 5500, pp. 2323--2326, 2000.

\bibitem{tenenbaum2000isomap}
J.~B. Tenenbaum, V.~d. Silva, and J.~C. Langford, ``A global geometric
  framework for nonlinear dimensionality reduction,'' \emph{Science}, vol. 290,
  no. 5500, pp. 2319--2323, 2000.

\bibitem{kruskal1964mds}
J.~B. Kruskal, ``Multidimensional scaling by optimizing goodness of fit to a
  nonmetric hypothesis,'' \emph{Psychometrika}, vol.~29, no.~1, pp. 1--27, Mar
  1964.

\bibitem{jolliffe2011pca}
I.~T. Jolliffe, ``Principal component analysis,'' in \emph{International
  Encyclopedia of Statistical Science}.\hskip 1em plus 0.5em minus 0.4em\relax
  Springer, 2011, pp. 1094--1096.

\bibitem{meng2011new}
D.~Meng, Y.~Leung, and Z.~Xu, ``A new quality assessment criterion for
  nonlinear dimensionality reduction,'' \emph{Neurocomputing}, vol.~74, no.~6,
  pp. 941--948, 2011.

\bibitem{cunningham2015linear}
J.~P. Cunningham and Z.~Ghahramani, ``Linear dimensionality reduction: survey,
  insights, and generalizations,'' \emph{J. Mach. Learn. Res.}, vol.~16, pp.
  2859--2900, 2015.

\bibitem{poli2008field}
R.~Poli, W.~B. Langdon, and N.~F. McPhee, \emph{A Field Guide to Genetic
  Programming}.\hskip 1em plus 0.5em minus 0.4em\relax lulu.com, 2008, (Last
  Accessed: 27/09/19).

\bibitem{lensen2019can}
A.~Lensen, B.~Xue, and M.~Zhang, ``Can genetic programming do manifold learning
  too?'' in \emph{Proceedings of the European Conference on Genetic Programming
  (EuroGP)}, ser. Lecture Notes in Computer Science, vol. 11451.\hskip 1em plus
  0.5em minus 0.4em\relax Springer, 2019, pp. 114--130.

\bibitem{orzechowski2020benchmarking}
P.~Orzechowski, F.~Magiera, and J.~H. Moore, ``Benchmarking manifold learning
  methods on a large collection of datasets,'' in \emph{Proceedings of the
  European Conference on Genetic Programming (EuroGP)}, 2020, pp. 135--150.

\bibitem{lensen2020genetic}
A.~Lensen, B.~Xue, and M.~Zhang, ``Genetic programming for evolving a front of
  interpretable models for data visualisation,'' \emph{{IEEE} Trans.
  Cybernetics}, pp. 1--15, February 2020, doi: 10.1109/TCYB.2020.2970198.

\bibitem{liu2012feature}
H.~Liu and H.~Motoda, \emph{Feature selection for knowledge discovery and data
  mining}.\hskip 1em plus 0.5em minus 0.4em\relax Springer Science \& Business
  Media, 2012, vol. 454.

\bibitem{xue2015survey}
B.~Xue, M.~Zhang, W.~N. Browne, and X.~Yao, ``A survey on evolutionary
  computation approaches to feature selection,'' \emph{{IEEE} Trans.
  Evolutionary Computation}, vol.~20, no.~4, pp. 606--626, 2016.

\bibitem{alsahaf2019survey}
H.~Al-Sahaf, Y.~Bi, Q.~Chen, A.~Lensen, Y.~Mei, Y.~Sun, B.~Tran, B.~Xue, and
  M.~Zhang, ``A survey on evolutionary machine learning,'' \emph{Journal of the
  Royal Society of New Zealand}, vol.~49, no.~2, pp. 205--228, April 2019.

\bibitem{espejo2010survey}
P.~G. Espejo, S.~Ventura, and F.~Herrera, ``A survey on the application of
  genetic programming to classification,'' \emph{{IEEE} Trans. Systems, Man,
  and Cybernetics, Part {C}}, vol.~40, no.~2, pp. 121--144, 2010.

\bibitem{neshatian2012filter}
K.~Neshatian, M.~Zhang, and P.~Andreae, ``A filter approach to multiple feature
  construction for symbolic learning classifiers using genetic programming,''
  \emph{{IEEE} Trans. Evolutionary Computation}, vol.~16, no.~5, pp. 645--661,
  2012.

\bibitem{lensen2017GPGC}
A.~Lensen, B.~Xue, and M.~Zhang, ``{GPGC:} genetic programming for automatic
  clustering using a flexible non-hyper-spherical graph-based approach,'' in
  \emph{Proceedings of the Genetic and Evolutionary Computation Conference,
  {GECCO}.}\hskip 1em plus 0.5em minus 0.4em\relax {ACM}, 2017, pp. 449--456.

\bibitem{lensen2019multi}
A.~Lensen, M.~Zhang, and B.~Xue, ``Multi-objective genetic programming for
  manifold learning: Balancing quality and dimensionality,'' \emph{Genetic
  Programming and Evolvable Machines}, vol.~21, pp. 399--431, 2020.

\bibitem{eppstein1997nn}
D.~Eppstein, M.~Paterson, and F.~F. Yao, ``On nearest-neighbor graphs,''
  \emph{Discret. Comput. Geom.}, vol.~17, no.~3, pp. 263--282, 1997.

\bibitem{chen2009local}
L.~Chen and A.~Buja, ``Local multidimensional scaling for nonlinear dimension
  reduction, graph drawing, and proximity analysis,'' \emph{Journal of the
  American Statistical Association}, vol. 104, no. 485, pp. 209--219, 2009.

\bibitem{venna2006local}
J.~Venna and S.~Kaski, ``Local multidimensional scaling,'' \emph{Neural
  Networks}, vol.~19, no. 6-7, pp. 889--899, 2006.

\bibitem{martins2014visual}
R.~M. Martins, D.~B. Coimbra, R.~Minghim, and A.~C. Telea, ``Visual analysis of
  dimensionality reduction quality for parameterized projections,''
  \emph{Comput. Graph.}, vol.~41, pp. 26--42, 2014.

\bibitem{aupetit2007visualizing}
M.~Aupetit, ``Visualizing distortions and recovering topology in continuous
  projection techniques,'' \emph{Neurocomputing}, vol.~70, no. 7-9, pp.
  1304--1330, 2007.

\bibitem{hinton2006reducing}
G.~E. Hinton and R.~R. Salakhutdinov, ``Reducing the dimensionality of data
  with neural networks,'' \emph{Science}, vol. 313, no. 5786, pp. 504--507,
  2006.

\bibitem{maaten2009learning}
L.~van~der Maaten, ``Learning a parametric embedding by preserving local
  structure,'' in \emph{Proceedings of the Twelfth International Conference on
  Artificial Intelligence and Statistics, {AISTATS}}, 2009, pp. 384--391.

\bibitem{yury2020efficient}
Y.~A. Malkov and D.~A. Yashunin, ``Efficient and robust approximate nearest
  neighbor search using hierarchical navigable small world graphs,''
  \emph{{IEEE} Trans. Pattern Anal. Mach. Intell.}, vol.~42, no.~4, pp.
  824--836, 2020.

\bibitem{aumuller2020ann}
M.~Aum{\"{u}}ller, E.~Bernhardsson, and A.~J. Faithfull, ``Ann-benchmarks: {A}
  benchmarking tool for approximate nearest neighbor algorithms,'' \emph{Inf.
  Syst.}, vol.~87, 2020.

\bibitem{saeed2018survey}
N.~Saeed, H.~Nam, M.~I.~U. Haq, and D.~M.~S. Bhatti, ``A survey on
  multidimensional scaling,'' \emph{{ACM} Comput. Surv.}, vol.~51, no.~3, pp.
  47:1--47:25, 2018.

\bibitem{amsaleg2015estimating}
L.~Amsaleg, O.~Chelly, T.~Furon, S.~Girard, M.~E. Houle, K.-i. Kawarabayashi,
  and M.~Nett, ``Estimating local intrinsic dimensionality,'' in
  \emph{Proceedings of the 21th ACM SIGKDD International Conference on
  Knowledge Discovery and Data Mining}, 2015, p. 29–38.

\bibitem{facco2017estimating}
E.~Facco, M.~d'Errico, A.~Rodriguez, and A.~Laio, ``Estimating the intrinsic
  dimension of datasets by a minimal neighborhood information,''
  \emph{Scientific Reports}, vol.~7, no.~1, p. 12140, Sep 2017.

\bibitem{uci}
\BIBentryALTinterwordspacing
D.~Dheeru and E.~Karra~Taniskidou, ``{UCI} machine learning repository,'' 2017.
  [Online]. Available: \url{http://archive.ics.uci.edu/ml}
\BIBentrySTDinterwordspacing

\bibitem{guyon2004result}
I.~Guyon, S.~R. Gunn, A.~Ben{-}Hur, and G.~Dror, ``Result analysis of the
  {NIPS} 2003 feature selection challenge,'' in \emph{Advances in Neural
  Information Processing Systems 17 [Neural Information Processing Systems,
  {NIPS} 2004, December 13-18, 2004, Vancouver, British Columbia, Canada]},
  2004, pp. 545--552.

\bibitem{yale2001}
A.~Georghiades, P.~Belhumeur, and D.~Kriegman, ``From few to many: Illumination
  cone models for face recognition under variable lighting and pose,''
  \emph{IEEE Trans. Pattern Anal. Mach. Intelligence}, vol.~23, no.~6, pp.
  643--660, 2001.

\bibitem{nene1996columbia}
S.~A. Nene, S.~K. Nayar, and H.~Murase, ``Columbia object image library
  (coil-20),'' Columbia University, Tech. Rep., 1996.

\bibitem{bentzien2013crowd}
J.~Bentzien, I.~Muegge, B.~Hamner, and D.~C. Thompson, ``Crowd computing: using
  competitive dynamics to develop and refine highly predictive models,''
  \emph{Drug Discovery Today}, vol.~18, no.~9, pp. 472--478, 2013.

\bibitem{alanni2019tumor}
R.~Alanni, J.~Hou, H.~Azzawi, and Y.~Xiang, ``A novel gene selection algorithm
  for cancer classification using microarray datasets,'' \emph{BMC Medical
  Genomics}, vol.~12, no.~1, p.~10, Jan 2019.

\bibitem{petricoin2002ovarian}
E.~F. Petricoin, A.~M. Ardekani, B.~A. Hitt, P.~J. Levine, V.~A. Fusaro, S.~M.
  Steinberg, G.~B. Mills, C.~Simone, D.~A. Fishman, E.~C. Kohn, and L.~A.
  Liotta, ``{{U}se of proteomic patterns in serum to identify ovarian
  cancer},'' \emph{Lancet}, vol. 359, no. 9306, pp. 572--577, Feb 2002.

\bibitem{scikit-learn}
F.~Pedregosa, G.~Varoquaux, A.~Gramfort, V.~Michel, B.~Thirion, O.~Grisel,
  M.~Blondel, P.~Prettenhofer, R.~Weiss, V.~Dubourg, J.~Vanderplas, A.~Passos,
  D.~Cournapeau, M.~Brucher, M.~Perrot, and E.~Duchesnay, ``Scikit-learn:
  Machine learning in {P}ython,'' \emph{Journal of Machine Learning Research},
  vol.~12, pp. 2825--2830, 2011.

\bibitem{tran2019genetic}
B.~Tran, B.~Xue, and M.~Zhang, ``Genetic programming for multiple-feature
  construction on high-dimensional classification,'' \emph{Pattern Recognit.},
  vol.~93, pp. 404--417, 2019.

\bibitem{chen2017feature}
Q.~Chen, M.~Zhang, and B.~Xue, ``Feature selection to improve generalization of
  genetic programming for high-dimensional symbolic regression,'' \emph{{IEEE}
  Trans. Evol. Comput.}, vol.~21, no.~5, pp. 792--806, 2017.

\bibitem{mei2017efficient}
Y.~Mei, S.~Nguyen, B.~Xue, and M.~Zhang, ``An efficient feature selection
  algorithm for evolving job shop scheduling rules with genetic programming,''
  \emph{{IEEE} Trans. Emerg. Top. Comput. Intell.}, vol.~1, no.~5, pp.
  339--353, 2017.

\end{thebibliography}
%

%
\begin{IEEEbiography}[{\includegraphics[width=1in,height=1.25in,clip,keepaspectratio]{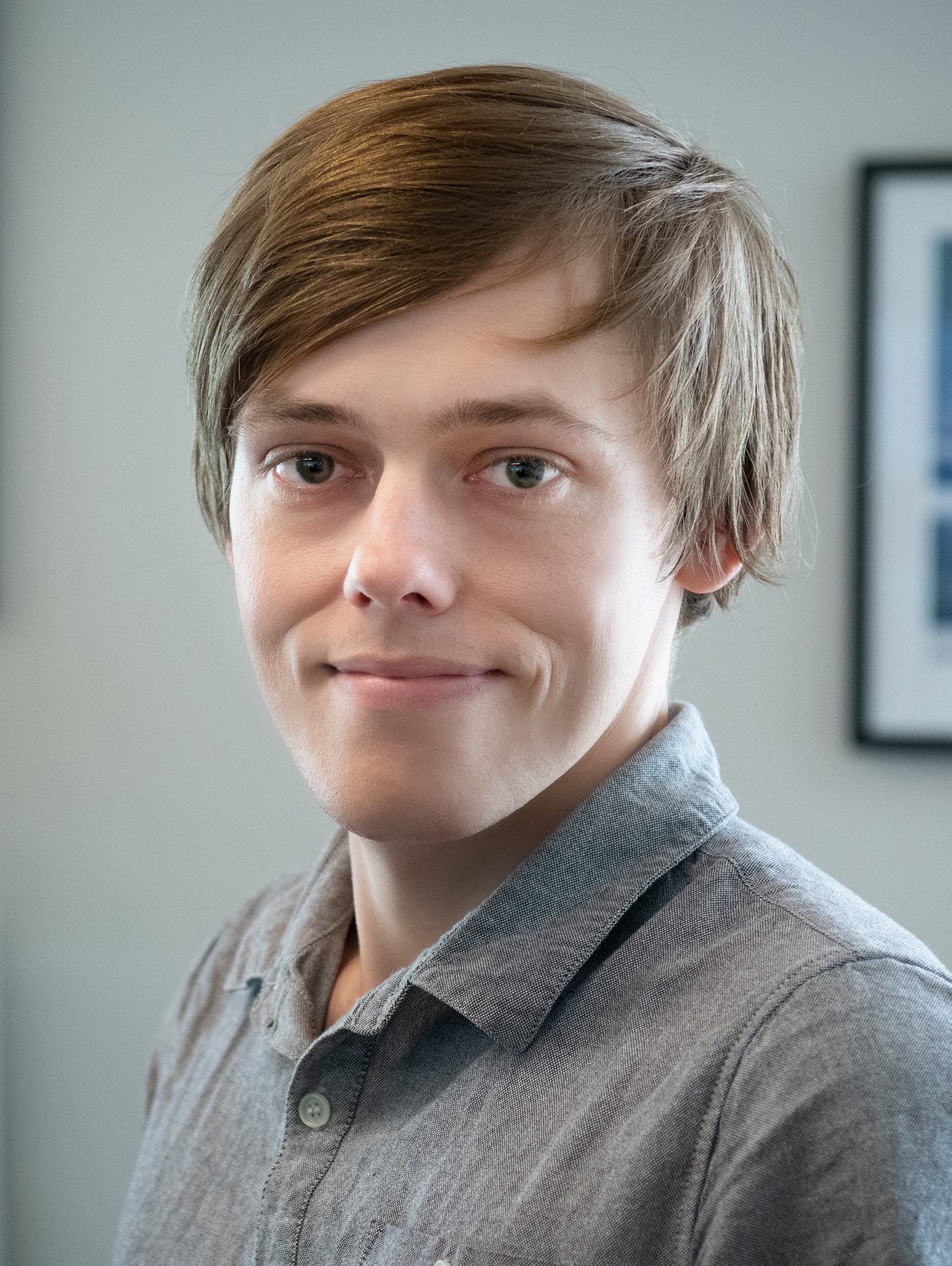}}]{Andrew Lensen} (M'17) received the B.Sc., B.Sc.(Hons 1$^{\text{st}}$ class), and Ph.D. degrees in computer science from Te Herenga Waka---Victoria University of Wellington (VUW), Wellington, New Zealand, in 2015, 2016, and 2019, respectively. 
	
He is currently a Lecturer in Artificial Intelligence (Pūkenga Atamai Horihori) in the Evolutionary Computation Research Group within the School of Engineering and Computer Science at Victoria University of Wellington. His current research interests are mainly in the use of evolutionary computation for feature manipulation in unsupervised learning, with a particular focus on the use of genetic programming for manifold learning and clustering, as well as explainable artificial intelligence (XAI) and AI ethics. 

Dr Lensen serves as a regular reviewer of several international conferences, including IEEE Congress on Evolutionary Computation, and international journals such as the IEEE Transactions on Evolutionary Computation, the IEEE Transactions on Cybernetics, and the IEEE Transactions on Emerging Topics in Computational Intelligence. He is also a committee member of the IEEE NZ Central Section and a member of the IEEE Computational Intelligence Society (CIS).
\end{IEEEbiography}
\begin{IEEEbiography}[{\includegraphics[width=1in,height=1.25in,clip,keepaspectratio]{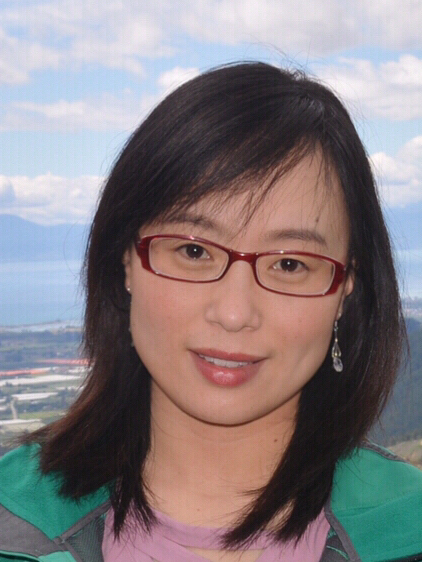}}]{Bing Xue} (M’10-SM'21) received the B.Sc.\ degree from the Henan University of Economics and Law, Zhengzhou, China, in 2007, the M.Sc.\ degree in management from Shenzhen University, Shenzhen, China, in 2010, and the Ph.D.\ degree in computer science in 2014 at VUW, New Zealand. 
	
She is currently a Professor in Computer Science, and Programme Director of Science in the School of Engineering and Computer Science at VUW. She has over 300 papers published in fully refereed international journals and conferences and her research focuses mainly on evolutionary computation, machine learning, neuroevolution, imagine analysis, transfer learning, and multi-objective machine learning.

Prof.\ Xue is the Chair of the IEEE CIS Task Force on Transfer Learning \& Transfer Optimization, Vice-Chair of the IEEE CIS Evolutionary Computation Technical Committee, Editor of IEEE CIS Newsletter, and Vice-Chair of IEEE Task Force on Evolutionary Feature Selection and Construction, and IEEE CIS Task Force on Evolutionary Deep Learning and Applications. She has served as associate editor of several international journals, such as IEEE Transactions on Evolutionary Computation and IEEE Transactions on Artificial Intelligence.
\end{IEEEbiography}
\begin{IEEEbiography}[{\includegraphics[width=1in,height=1.25in,clip,keepaspectratio]{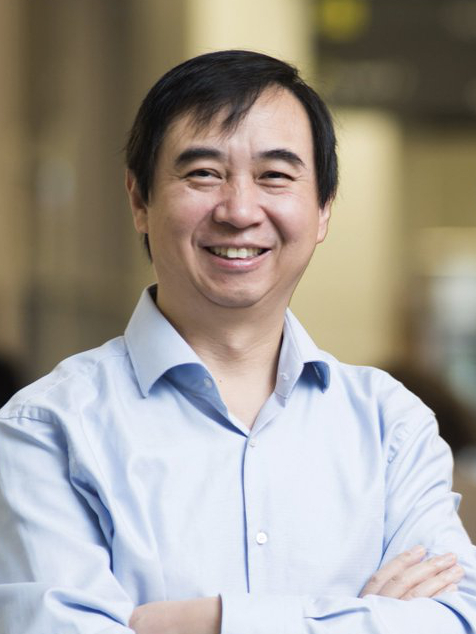}}]{Mengjie Zhang} (M’04–SM’10-F'19) received the B.E.\ and M.E.\ degrees from the Artificial Intelligence Research Center, Agricultural University of Hebei, Hebei, China, and the Ph.D.\ degree in computer science from RMIT University, Melbourne, Australia, in 1989, 1992, and 2000, respectively. 
	
He is currently Professor of Computer Science, Head of the Evolutionary Computation Research Group, and the Associate Dean (Research and Innovation) in the Faculty of Engineering, VUW. His current research interests include evolutionary computation, particularly genetic programming and particle swarm optimization, with application areas of image analysis, multi-objective optimization, feature selection and reduction, job shop scheduling, and transfer learning. He has published over 700 papers in refereed international journals and conferences. 
	
Prof.\ Zhang is a Fellow of the Royal Society of New Zealand, a Fellow of the IEEE, an IEEE CIS Distinguished Lecturer, and has been a Panel member of the Marsden Fund (New Zealand Government Funding). He was the chair of the IEEE CIS Intelligent Systems and Applications Technical Committee, the chair for the IEEE CIS Emergent Technologies Technical Committee and the Evolutionary Computation Technical Committee, and a member of the IEEE CIS Award Committee. He is a vice-chair of the IEEE CIS Task Force on Evolutionary Feature Selection and Construction, a vice-chair of the Task Force on Evolutionary Computer Vision and Image Processing, and the founding chair of the IEEE Computational Intelligence Chapter in New Zealand. He is also a committee member of the IEEE NZ Central Section.
\end{IEEEbiography}






\end{document}